# Normalised Clustering Accuracy: An Asymmetric External Cluster Validity Measure

**Marek Gagolewski**[1,2]

## Abstract

There is no, nor will there ever be, single best clustering algorithm. Nevertheless, we would still like to be able to distinguish between methods that work well on certain task types and those that systematically underperform. Clustering algorithms are traditionally evaluated using either internal or external validity measures. Internal measures quantify different aspects of the obtained partitions, e.g., the average degree of cluster compactness or point separability. However, their validity is questionable because the clusterings they endorse can sometimes be meaningless. External measures, on the other hand, compare the algorithms' outputs to fixed ground truth groupings provided by experts. In this paper, we argue that the commonly used classical partition similarity scores, such as the normalised mutual information, Fowlkes–Mallows, or adjusted Rand index, miss some desirable properties. In particular, they do not identify worst-case scenarios correctly, nor are they easily interpretable. As a consequence, the evaluation of clustering algorithms on diverse benchmark datasets can be difficult. To remedy these issues, we propose and analyse a new measure: a version of the optimal set-matching accuracy, which is normalised, monotonic with respect to some similarity relation, scale-invariant, and corrected for the imbalancedness of cluster sizes (but neither symmetric nor adjusted for chance).

## 1 Introduction

Clustering is an unsupervised learning technique that aims at identifying semantically useful partitions of a given dataset (Hennig, 2015; von Luxburg et al., 2012). Up to this date, many clustering algorithms have been proposed, but the problem of how to evaluate the overall quality of the outputs they generate is still open for discussion (Xiong & Li, 2014; Tavakkol et al., 2022; Ullmann et al., 2022; van Mechelen et al., 2023). We know that there will never be a single "best" method (Ackerman et al., 2021; Strobl & Leisch, 2022). Nevertheless, at the very least, we would like to be able to identify the algorithms that are somewhat sensible on certain classes of datasets, or filter out those that consistently yield disappointing results.

✉ Marek Gagolewski
marek.gagolewski@pw.edu.pl

[1] Faculty of Mathematics and Information Science, Warsaw University of Technology, ul. Koszykowa 75, 00-662 Warsaw, Poland

[2] Systems Research Institute, Polish Academy of Sciences, ul. Newelska 6, 01-447 Warsaw, Poland

*Internal validity measures*, such as the Caliński–Harabasz, Dunn, or Silhouette index (Caliński & Harabasz, 1974; Dunn, 1974; Rousseeuw, 1987), are often used to quantify how well a given partition reflects the structure of the underlying unknown data distribution, for instance, the degree of compactness or separability (e.g., Milligan and Cooper, 1985; Maulik and Bandyopadhyay, 2002; Halkidi et al., 2001; Arbelaitz et al., 2013; Xu et al., 2020). However, they have been recently criticised by Gagolewski et al. (2021) who noted that some popular measures often promote clusterings that are not meaningful, e.g., they return cluster memberships that resemble noise or should rather be employed as outlier detectors (see Fig. 1 therein).

*External validity measures*, on the other hand, operate under the assumption that benchmark datasets are equipped with expert-given labels; this is true for many recently introduced test batteries (Graves & Pedrycz, 2010; Thrun & Ultsch, 2020; Dua & Graff, 2022; Fränti & Sieranoja, 2018; Gagolewski, 2022). Moreover, they presume that it would be best if an algorithm returned a clustering that is as similar to the reference one as possible. In the literature, it has become customary to use *partition similarity scores* as external measures, e.g., the adjusted Rand, normalised mutual information, or pair sets indices (Wagner & Wagner, 2006; Horta & Campello, 2015; Rezaei & Fränti, 2016). However, in this paper, we argue that simpler objects can actually be more suitable.

Let $X_1, \ldots, X_k$ be a *reference* (ground truth, indicated by experts) partition of a set $X$ of $n$ objects into $k$ nonempty and pairwise disjoint clusters. Moreover, let $\hat{X}_1, \ldots, \hat{X}_k$ be a *predicted* (generated by a clustering algorithm) partition of $X$ into $k$ disjoint clusters. Commonly, the knowledge about which clusters correspond to one another is summarised by a $k \times k$ confusion matrix $\mathbf{C} = (c_{i,j})$ whose entry in the $i$-th row and the $j$-th column gives the number of elements in the $i$-th true cluster that an algorithm allocated to the $j$-th predicted cluster, i.e., $c_{i,j} = \#(X_i \cap \hat{X}_j)$.

In this paper, we are interested in studying real-valued functions aiming at quantifying how similar a predicted partition is to the reference one; compare Fig. 1. For a measure to be useful in the task at hand, it should meet a number of desirable properties (postulates). For the purpose of this introduction, we will now state them only descriptively; the formalism will follow in Sect. 3:

- [MON] The more *similar* the partitions are, the higher the score should be (in our case, we will consider monotonicity with respect to the diagonal max-dominance relation that we define below).
- [B1] Only two identical partitions yield the highest possible similarity score, which we conventionally assume to be equal to 1.
- [PER] The score does not change if we relabel the cluster IDs (i.e., swapping $X_i \leftrightarrow X_j$ or $\hat{X}_i \leftrightarrow \hat{X}_j$ for any $i \neq j$; this comports with a rearrangement of rows or columns in the corresponding confusion matrix).

The third property stems from the fact that a partition is a set of clusters, and sets (equivalence classes in the equivalence relation "two points belong to the same cluster") are unordered by definition. From this perspective, "Gaussian mixture" and "Gaussian mixture relabelled" in Fig. 1 represent identical partitions. Overall, clustering is an unsupervised learning problem; therefore, we cannot expect an algorithm to guess the order of reference labels.

**Remark 1** Traditional approaches to defining classes of partition similarity scores usually require the fulfilment of the above [MON], [B1], and [PER]. However, additionally (see,

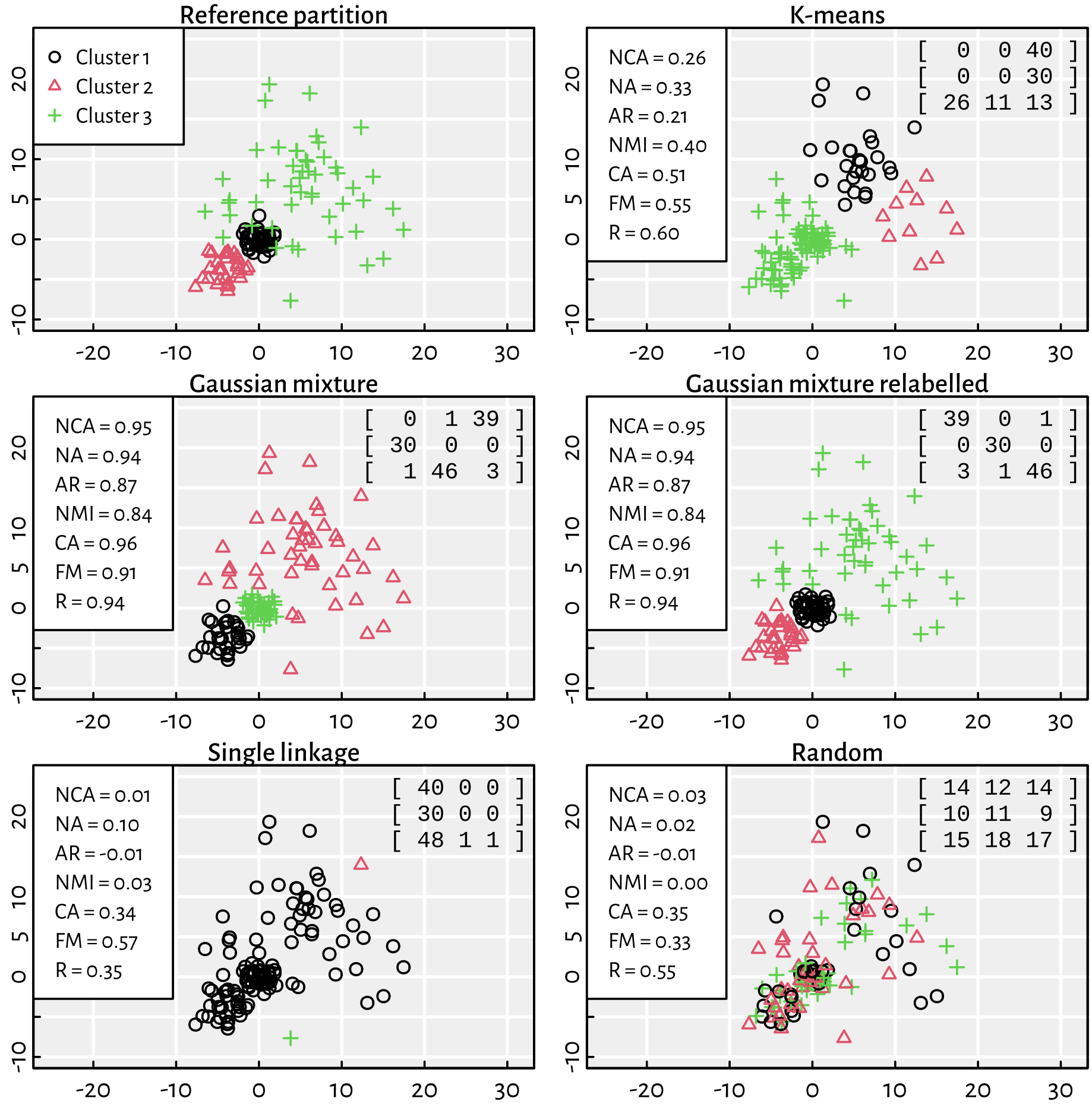

**Fig. 1** A reference (ground truth; $k = 3$) partition of an example dataset (WUT/x2; $n = 120$; see Sect. 4) and some predicted clusterings that we would like to relate to it. We also report the confusion matrices and the values of a few external cluster validity measures defined in Sects. 2 and 3. The Gaussian mixture algorithm misclassifies only five points (ca 4%), but the indices' values are quite different. In our case, the non-normalised measures (here: CA, FM, and R) do not distinguish between the cases of two undesirable partition types: assigning most points into a single cluster (as returned by single linkage) and memberships assigned at random. Note that there can be many possible reference partitions for a given dataset (Dasgupta & Ng, 2009; Gagolewski, 2022): the clusterings returned both by the Gaussian mixture method and K-Means can be considered meaningful (at least, in the current author's opinion). However, we are only relating a predicted partition to one reference clustering at a time

e.g., Meilă, 2005; Wagner and Wagner, 2006; Xiong and Li, 2014; Rezaei and Fränti, 2016; Arinik et al. 2021) they assume that:

- [SYM] The score is the same if we swap the roles of the two partitions.

In other words, none of the two partitions is treated specially. However, in the context of our paper, we consider $X_1, \ldots, X_k$ as a fixed point of reference for $\hat{X}_1, \ldots, \hat{X}_k$. Therefore, requiring this condition is not necessary.

Further postulates are related to the worst-case scenarios. When evaluating clustering algorithms on many diverse datasets, we report aggregated similarity scores. It may thus be desirable to have the lower index bound not dependent on, amongst others, the number of clusters $k$, which may vary across benchmark instances. In other words, we may[1] want to have the indices on the same scale: a common choice is the unit interval. And thus, in Sect. 3, we will discuss the following properties.

- [B0] The smallest possible value of the index is 0.
- [U0] Perfectly uniform assignment of points to predicted clusters, i.e., when $c_{i,j} = c_{i,i}$ for all $i, j$, results in the similarity score of 0.
- [O0] Assigning all the points to a single cluster gives the similarity score of 0.

We will also relate the above to the adjustment for chance, where the *expected* value of an index is 0 given two independent partitions having the same marginal frequencies (property [E0]). Nevertheless, we will note that, in general, it contradicts the three above properties.

Moreover, we will also be interested in the following types of scale invariances.

- [SU] The similarity score should remain the same when we double, triple, etc., the number of points in each subset $X_i \cap \hat{X}_j$, without changing the structure of the discovered clusters (the output for $s\mathbf{C}$ should be the same as the one for $\mathbf{C}$ for any $s > 0$).
- [SC] The similarity score should be the same if we multiply points in one reference cluster. In particular, it should not be affected by the imbalancedness of the true cluster sizes, whether some clusters are relatively small or large (the output for $\mathbf{C}$ should be the same as $\mathrm{diag}(s_1, \ldots, s_k)\,\mathbf{C}$ for any $s_1, \ldots, s_k > 0$).

The paper is structured as follows. In the next section, we introduce three basic classes of partition similarity scores. In Sect. 3, we formalise the above properties, study which indices fulfil them, and discuss their various modifications. In particular, we derive a new score called *normalised clustering accuracy*, which is given by:

$$\mathrm{NCA}(\mathbf{C}) = \max_{\sigma:\{1,\ldots,k\} \overset{\text{biject.}}{\to} \{1,\ldots,k\}} \frac{1}{k} \sum_{i=1}^{k} \frac{c_{i,\sigma(i)} - \frac{1}{k} c_{i,\cdot}}{c_{i,\cdot} - \frac{1}{k} c_{i,\cdot}}, \tag{1}$$

where $c_{i,\cdot}$ is the number of elements in the $i$-th true cluster.

NCA is the averaged percentage of correctly classified points in each cluster *above* the perfectly uniform cluster membership assignment. NCA relies on finding the permutation (relabelling) that gives the optimal matching of cluster IDs between the reference and the predicted set, which can be computed in $O(k^3)$ time. We will prove that it is normalised to the unit interval, monotonic with respect to a particular similarity relation, and corrected for cluster sizes' imbalancedness.

Further, in Sect. 4, we analyse the degree of association between pairs of indices on an example benchmark dataset battery and inspect how the choice of a similarity measure affects the rankings of a few well-known clustering algorithms. Section 5 sketches possible extensions of the introduced measure to the case of clusterings of different cardinalities.

The implementations of the discussed measures are included in the open-source *genieclust* (Gagolewski, 2021) package for Python and R.

[1] One of the reviewers of this paper noted that "[B0] is as much a convention (rather than a desirable feature) as the maximum of 1 is. Allowing for negative values is not a reason against adjusted Rand for example. In fact, [B0] implies that an index cannot have the expected value 0 on random partitions, and the latter can be seen as desirable (later covered as [E0])." We acknowledge that in certain applications, [E0] might indeed be more desirable than [B0].

## 2 External Cluster Validity Measures

Let $\mathbf{C} = (c_{i,j})$ be a *confusion (matching) matrix* of size $k \times k$, whose rows correspond to the reference clusters, and columns summarise the predicted cluster memberships:

$$\mathbf{C} = \begin{array}{c} \left[\begin{array}{cccc} c_{1,1} & c_{1,2} & \cdots & c_{1,k} \\ c_{2,1} & c_{2,2} & \cdots & c_{2,k} \\ c_{3,1} & c_{3,2} & \cdots & c_{3,k} \\ \vdots & \vdots & \ddots & \vdots \\ c_{k,1} & c_{k,2} & \cdots & c_{k,k} \end{array}\right] \begin{array}{c} c_{1,\cdot} \\ c_{2,\cdot} \\ c_{3,\cdot} \\ \vdots \\ c_{k,\cdot} \end{array} \\ \begin{array}{cccc} c_{\cdot,1} & c_{\cdot,2} & \cdots & c_{\cdot,k} \end{array} \quad n \end{array}$$

For brevity, we denote the total sum by $n = \sum_{i=1}^{k} \sum_{j=1}^{k} c_{i,j}$, and the row- and column-wise sums by $c_{i,\cdot} = \sum_{j=1}^{k} c_{i,j}$ and $c_{\cdot,j} = \sum_{i=1}^{k} c_{i,j}$.

In the classical (crisp) clustering setting, all $c_{i,j}$s are nonnegative integers.[2] In such a case, $c_{i,j}$ denotes the number of points from the $i$-th reference cluster that an algorithm assigns to the $j$-th predicted group. Moreover, then $n$ gives the total number of points, $c_{i,\cdot}$ is the cardinality of the $i$-th reference cluster, and $c_{\cdot,j}$ is the size of the $j$-th predicted one. Here, the clusterings can be represented by means of two label vectors $\mathbf{y} = (y_u)$ and $\hat{\mathbf{y}} = (\hat{y}_u)$ of length $n$, where $y_u, \hat{y}_u \in \{1, \ldots, k\}$ denote the true and predicted memberships of the $u$-th point. We thus have $c_{i,j} = \#\{u : y_u = i \text{ and } \hat{y}_u = j\}$, $c_{i,\cdot} = \#\{u : y_u = i\}$, and $c_{\cdot,j} = \#\{u : \hat{y}_u = j\}$.

Let us review some of the most seminal classes of crisp partition similarity scores to which we will relate our proposal. More measures are discussed by, amongst others, Xiong and Li (2014); Wagner and Wagner (2006); Rezaei and Fränti (2016); Arinik et al. (2021).

### 2.1 Counting Concordant and Discordant Point Pairs

The first class of indices is based on counting point pairs that are concordant:

- $\mathcal{YY} = \#\left\{i < j : y_i = y_j \text{ and } \hat{y}_i = \hat{y}_j\right\} = \sum_{i=1}^{k} \sum_{j=1}^{k} \binom{c_{i,j}}{2}$,
- $\mathcal{NN} = \#\left\{i < j : y_i \neq y_j \text{ and } \hat{y}_i \neq \hat{y}_j\right\} = \binom{n}{2} - \mathcal{YY} - \mathcal{NY} - \mathcal{YN}$,

and those that are discordant:

- $\mathcal{NY} = \#\left\{i < j : y_i \neq y_j \text{ but } \hat{y}_i = \hat{y}_j\right\} = \sum_{i=1}^{k} \binom{c_{i,\cdot}}{2} \mathcal{YY}$,
- $\mathcal{YN} = \#\left\{i < j : y_i = y_j \text{ but } \hat{y}_i \neq \hat{y}_j\right\} = \sum_{j=1}^{k} \binom{c_{\cdot,j}}{2} \mathcal{YY}$;

see the paper by Hubert and Arabie (1985) for discussion.

In particular, the *Rand index* (Rand, 1971) is defined as the classification accuracy:

$$\mathrm{R}(\mathbf{C}) = \frac{\mathcal{YY} + \mathcal{NN}}{\binom{n}{2}} = 1 - \frac{\sum_{i=1}^{k} \binom{c_{i,\cdot}}{2} + \sum_{j=1}^{k} \binom{c_{\cdot,j}}{2} - 2 \sum_{i=1}^{k} \sum_{j=1}^{k} \binom{c_{i,j}}{2}}{\binom{n}{2}} \tag{2}$$

$$= 1 - \frac{\sum_{i=1}^{k} \left(c_{i,\cdot}^2 - \sum_{j=1}^{k} c_{i,j}^2\right) + \sum_{j=1}^{k} \left(c_{\cdot,j}^2 - \sum_{i=1}^{k} c_{i,j}^2\right)}{n(n-1)}. \tag{3}$$

[2] We shall relax this condition below. Furthermore, the more complicated case where the number of reference and predicted clusters might differ will also be considered.

Moreover, the *Fowlkes–Mallows index* (Fowlkes & Mallows, 1983) is the geometric mean between precision and recall:

$$\mathrm{FM}(\mathbf{C}) = \frac{\mathcal{YY}}{\sqrt{(\mathcal{YY}+\mathcal{YN})(\mathcal{YY}+\mathcal{NY})}} = \frac{\sum_{i=1}^{k}\sum_{j=1}^{k}\binom{c_{i,j}}{2}}{\sqrt{\sum_{i=1}^{k}\binom{c_{i,\cdot}}{2}}\sqrt{\sum_{j=1}^{k}\binom{c_{\cdot,j}}{2}}} \tag{4}$$

$$= \frac{\sum_{i=1}^{k}\sum_{j=1}^{k} c_{i,j}^2 - n}{\sqrt{\sum_{i=1}^{k} c_{i,\cdot}^2 - n}\sqrt{\sum_{j=1}^{k} c_{\cdot,j}^2 - n}}. \tag{5}$$

A detailed overview of the behaviour of the indices based on counting object pairs is presented by Warrens and van der Hoef (2022); Horta and Campello (2015); Lei et al. (2017). In the sequel, we will consider various variations of the above indices, e.g., their normalised and adjusted for chance versions.

### 2.2 Information-Theoretic Measures

Another group of partition similarity scores consists of the so-called information-theoretic measures. As a point of departure for further derivations, let us recall the *mutual information* score (Horibe, 1985):

$$\mathrm{MI}(\mathbf{C}) = \sum_{i=1}^{k}\sum_{j=1}^{k} \frac{c_{i,j}}{n} \log \frac{n\, c_{i,j}}{c_{i,\cdot}\, c_{\cdot,j}}, \tag{6}$$

with convention $0 \log x = 0$ for any $x \geq 0$. Two variants of this index will be presented below; for further discussion, see the papers by Vinh et al. (2010); van der Hoef and Warrens (2019).

### 2.3 Accuracy-like Set-matching Measures

Let us note that the Rand and the Fowlkes–Mallows scores use $1/\binom{n}{2}$ as the unit of information. Sometimes, we might prefer working on the $1/n$-based scale. However, it would be a mistake to rely on the standard accuracy as known from the evaluation of classification models:

$$\ddot{\mathrm{A}}(\mathbf{C}) = \sum_{i=1}^{k} \frac{c_{i,i}}{n}, \tag{7}$$

which is the proportion of correctly classified points. This measure should not be used in clustering because clusters are defined up to a permutation of the sets' IDs (compare the desired property [PER]).

In our context, the predicted clusters need to be matched with the true ones somehow. For instance, the confusion matrix corresponding to the label vectors $\mathbf{y} = (1, 2, 2, 1, 2, 3, 1, 1, 1)$ and $\hat{\mathbf{y}} = (3, 1, 1, 3, 1, 2, 3, 3, 3)$,

$$\mathbf{C} = \begin{bmatrix} 0 & 0 & 5 \\ 3 & 0 & 0 \\ 0 & 1 & 0 \end{bmatrix}, \tag{8}$$

represents a perfect match. Hence, in this case, we could use $(c_{2,1} + c_{3,2} + c_{1,3})/n$ as a measure of accuracy. Consequently, we need an algorithmic way to translate between the predicted and reference cluster IDs. The simplest choice involves greedy pairing:

$$\ddot{\mathrm{P}}(\mathbf{C}) = \sum_{j=1}^{k} \max_{i=1}^{k} \frac{c_{i,j}}{n}, \tag{9}$$

or:

$$\ddot{\mathrm{P}}^T(\mathbf{C}) = \ddot{\mathrm{P}}(\mathbf{C}^T) = \sum_{i=1}^{k} \max_{j=1}^{k} \frac{c_{i,j}}{n}. \tag{10}$$

These measures are sometimes referred to as *purity*. Unfortunately, there is no guarantee that all clusters will welcome a match.

If we want to remedy this issue, we should seek a one-to-one correspondence between the cluster IDs, $\sigma$, which is a solution to the following optimisation problem:

$$\text{maximise} \sum_{i=1}^{k} c_{i,\sigma(i)} \qquad \text{w.r.t. } \sigma \in \mathfrak{S}_k, \tag{11}$$

where $\mathfrak{S}_k$ is the set of all permutations of the set $\{1, \ldots, k\}$, i.e., bijections from $\{1, \ldots, k\}$ to itself.

The above guarantees that each column is paired with one and only one row in the confusion matrix. Such optimal pairing leads to, what we call here, the *pivoted accuracy* (Meilă and Heckerman, 2001, Eq. (13); Steinley, 2004, Eq. (10); Charon et al., 2006; Chacón, 2021):

$$\mathrm{A}(\mathbf{C}) = \max_{\sigma \in \mathfrak{S}_k} \sum_{i=1}^{k} \frac{c_{i,\sigma(i)}}{n}. \tag{12}$$

It relies on the best matching of the labels in the reference partition to the labels in the predicted grouping so as to maximise the standard accuracy. At first glance, it is a very attractive measure because of its interpretability (a feature whose importance was already noted by Goodman and Kruskal (1979)). However, soon, we will note that its values for the worst possible clusterings depend on the number of clusters $k$.

**Remark 2** Eq. 11 can be expressed as the following 0–1 integer linear programming problem:

$$\text{maximise} \sum_{j=1}^{k} \sum_{i=1}^{k} b_{i,j} c_{i,j} \qquad \text{w.r.t. } \mathbf{B} \in \{0, 1\}^{k \times k}, \tag{13}$$

such that $\mathbf{B} = (b_{i,j})$ is a binary matrix with one and only one value 1 in every row and in every column, i.e., $b_{i,\cdot} = 1$ and $b_{\cdot,j} = 1$ for all $i, j$. It is called the *maximal linear sum assignment problem* or maximum bipartite graph matching. It can be solved using, e.g., the so-called Hungarian algorithm, which requires $O(k^3)$ time (Crouse, 2016).

**Remark 3** Note that optimal matching is not the same as the greedy recursive pairing, where we pick the largest element, and then continue the same procedure in the yet-to-be-selected parts of the matrix. For example, given:

$$\mathbf{C} = \begin{bmatrix} 50 & 25 & 25 \\ 21 & 40 & 39 \\ 39 & 39 & 22 \end{bmatrix}, \tag{14}$$

the optimal permutation is $\sigma = (1, 3, 2)$, yielding the sum $50 + 39 + 39 = 128$, whereas the greedy recursive pairing gives $\sigma = (1, 2, 3)$, corresponding to $50 + 40 + 22 = 112$.

In what follows, we will discuss many modifications of the aforementioned indices.

## 3 Desirable Properties and Features of Indices

Let $\mathfrak{C}^{k\times k}$ denote the set of admissible confusion matrices of size $k \times k$ such that if $\mathbf{C} = (c_{i,j}) \in \mathfrak{C}^{k\times k}$, then $c_{i,j} \geq 0$ and $c_{i,\cdot} > 0$ for all $i, j$. Note that the case where the confusion matrix is non-square will be mentioned separately later (Sect. 5).

Even though we assume that our *reference* partitions are crisp, in general, we do not want to restrict ourselves to classical (crisp) clustering only. Thus, whether we allow $c_{i,j}$s to be arbitrary nonnegative real numbers (not just integer ones), will depend on the index. This will not only simplify the further analysis, but also allow us to accommodate, amongst others, the case of weighted (fuzzy) *predicted* clusterings, where point memberships are described by probability vectors; compare, e.g., the papers by Hüllermeier et al. (2012); Campagner et al. (2023); Andrews et al. (2022); Flynt et al. (2019); D'Ambrosio et al. (2021); Horta and Campello (2015). We will see that under the scale invariance [SU] property discussed below, this will come without loss in generality.

Additionally, we assume that none of the reference clusters is empty. However, we actually allow a clustering algorithm to return a partition of lower cardinality than $k$, which is represented by the case of some $c_{\cdot,j}$'s being equal to 0.

In Sect. 1, we outlined a few desirable properties of cluster validity measures in a rather informal manner. Let us formalise them now so that we can introduce various adjustments to the indices. For brevity, all the definitions below should be read as "We say that an index I meets Property [X], whenever for any $k \geq 2$ and…". A summary will be given in Table 2.

### 3.1 Permutation Invariance

For any $\sigma \in \mathfrak{S}_k$, denote by $\mathbf{P}_\sigma = (p_{i,j})$ the corresponding permutation matrix, i.e., a $k \times k$ binary matrix with $p_{i,j} = 1$ if and only if $i = \sigma(j)$. We stated that clusters are defined up to a permutation of cluster IDs. Therefore, rearranging rows or columns of a confusion matrix should not affect the index value.

**Definition 1** (Property [PER]) For all $\mathbf{C} \in \mathfrak{C}^{k\times k}$ and every permutation $\sigma \in \mathfrak{S}_k$, we have $\mathrm{I}(\mathbf{C}) = \mathrm{I}(\mathbf{P}_\sigma \mathbf{C}) = \mathrm{I}(\mathbf{C}\mathbf{P}_\sigma)$.

As the standard classification accuracy $\ddot{\mathrm{A}}$ (7) is the only index amongst the ones considered in this paper that does not fulfil this property, we will no longer be considering it a viable cluster validity measure candidate.

### 3.2 Symmetry

In traditional partition similarity scores, both partitions should be treated equally. This is reflected by the symmetry property.

**Definition 2** (Property [SYM]) For all $\mathbf{C} \in \mathfrak{C}^{k\times k}$, we have $\mathrm{I}(\mathbf{C}) = \mathrm{I}(\mathbf{C}^T)$.

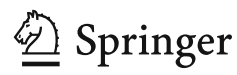

Disregarding from now on the two versions of purity $\ddot{\mathrm{P}}$ and $\ddot{\mathrm{P}}^T$, which are semantically problematic, all the aforementioned scores enjoy this condition. However, we have already argued that in the context of external cluster validation, [SYM] can be omitted. For a given benchmark problem, the reference partition is fixed. Thus, we treat it differently from the predicted ones, because the latter vary across the algorithms under scrutiny.

### 3.3 Scale Invariance

Another property we might find useful is that any scaling of the confusion matrix should not change the similarity assessment. Scaling can be interpreted as adding or removing new points to the detected clusters without disturbing the discovered structure while maintaining the proportions of cluster sizes. In other words, if we have 50% points correctly classified, whether this was achieved for $n = 100$ or $n = 10{,}000$ should not matter.

**Definition 3** (Property [SU]) For all $\mathbf{C} \in \mathfrak{C}^{k\times k}$ and every $s > 0$, we have $\mathrm{I}(s\mathbf{C}) = \mathrm{I}(\mathbf{C})$.

Amongst the indices studied so far, only those based on counting concordant/discordant point pairs do not enjoy this property.

**Example 1** For instance, for $\mathbf{C}$ given by Eq. 14, we have $\mathrm{FM}(\mathbf{C}) \approx 0.35297$ but $\mathrm{FM}(3\mathbf{C}) \approx 0.35727$, and $\mathrm{R}(\mathbf{C}) \approx 0.56928$ but $\mathrm{R}(3\mathbf{C}) \approx 0.57023$. The difference becomes even more significant when we consider non-integer confusion matrices. Assuming that we generalise $\binom{x}{2}$ for arbitrary reals as $x(x-1)/2$, we get, e.g., $\mathrm{FM}(\frac{1}{900}\mathbf{C}) \approx -1.08054$ and $\mathrm{R}(\frac{1}{900}\mathbf{C}) \approx 1.21464$. This is why, in the context of R and FM scores, the study of their properties will be limited to integer matrices only.

However, we can consider $\mathrm{R}'(\mathbf{C}) = \lim_{s\to\infty} \mathrm{R}(s\mathbf{C})$ and $\mathrm{FM}'(\mathbf{C}) = \lim_{s\to\infty} \mathrm{FM}(s\mathbf{C})$. Noting that in Eqs. 3 and 5, if the total sum of $\mathbf{C}$ is $n$, then the sum of $s\mathbf{C}$ is $sn$, these limits are given by:

$$\mathrm{R}'(\mathbf{C}) = 1 - \frac{\sum_{i=1}^{k}\left(c_{i,\cdot}^2 - \sum_{j=1}^{k} c_{i,j}^2\right) + \sum_{j=1}^{k}\left(c_{\cdot,j}^2 - \sum_{i=1}^{k} c_{i,j}^2\right)}{n^2}, \tag{15}$$

$$\mathrm{FM}'(\mathbf{C}) = \frac{\sum_{i=1}^{k}\sum_{j=1}^{k} c_{i,j}^2}{\sqrt{\sum_{i=1}^{k} c_{i,\cdot}^2}\sqrt{\sum_{j=1}^{k} c_{\cdot,j}^2}}. \tag{16}$$

The above arise by replacing $\binom{x}{2}$ with $x^2/2$ in Eqs. 2 and 4. Therefore, the price for the fulfilment of the scale invariance property is the loss of the original interpretation related to the counting of concordant pairs.

### 3.4 Upper Bound

From now on, let $\mathfrak{C}^{k\times k}_{s_1,\dots,s_k} \subseteq \mathfrak{C}^{k\times k}$ denote the set of confusion matrices like $\mathbf{C} = (c_{i,j})$ whose $i$-th row's sum $c_{i,\cdot}$ is equal to $s_i$, for all $i = 1, \dots, k$.

In order for an index value to be interpretable, we need to calibrate it so that we know which scores indicate low and high-quality outcomes. In the literature, it is customary to assign the value of 1 when there is a perfect match, and preferably only then. In clustering problems, it is the case when each predicted set can be mapped to precisely one reference cluster, and vice versa. This is represented by confusion matrices with 0 s everywhere but on the diagonal or (by [PER]) their permuted versions; compare, e.g., Eq. 8.

**Definition 4** (Property [B1]) For all $s_1, \ldots, s_k > 0$ and every $\mathbf{C} \in \mathfrak{C}^{k\times k}_{s_1,\ldots,s_k}$, we have $\mathrm{I}(\mathbf{C}) \le 1$. Moreover, $\mathrm{I}(\mathbf{C}) = 1$ if and only if there exists a permutation $\sigma \in \mathfrak{S}_k$ such that $\mathbf{C} = \mathbf{P}_\sigma \mathbf{S}$, where $\mathbf{S} = \mathrm{diag}(s_1, \ldots, s_k)$.

The Rand and Fowlkes-Mallows indices enjoy [B1] when restricted to the case of integer matrices. Mutual information is not normalised; e.g., for a matrix like $\mathbf{I} = \mathrm{diag}(s, \ldots, s)$, we have $\mathrm{MI}(\mathbf{I}) = \log k$ for all $s > 0$. Other indices discussed so far fulfil this property.

**Example 2** The MI score can be rescaled so as to meet [B1]. For instance, the *normalised mutual information* score denoted by $I/D_2$ by Kvalseth (1987) and $\mathrm{NMI}_{\mathrm{sum}}$ by Vinh et al. (2010) is given by:

$$\mathrm{NMI}(\mathbf{C}) = \frac{\sum_{i=1}^{k}\sum_{j=1}^{k} \frac{c_{i,j}}{n} \log \frac{n c_{i,j}}{c_{i,\cdot} c_{\cdot,j}}}{\frac{1}{2}\left(\sum_{i=1}^{k} \frac{c_{i,\cdot}}{n} \log \frac{n}{c_{i,\cdot}} + \sum_{j=1}^{k} \frac{c_{\cdot,j}}{n} \log \frac{n}{c_{\cdot,j}}\right)}. \tag{17}$$

It meets [PER], [SYM], [SU], and [B1].

### 3.5 Adjustment for Chance

In the context of comparing partitions, some statistical literature finds it desirable if two groupings generated independently at random are assigned a similarity score of 0 *on average* (Hubert & Arabie, 1985; Vinh et al., 2010).

For instance, under the hypergeometric model for randomness discussed by Fowlkes and Mallows (1983) (for possible alternatives, see Steinley (2004) and Gates and Ahn (2017)), the two partitions are assumed to be picked independently at random subject to having the original $n$, $k$, and counts of objects in each cluster, i.e., $c_{1,\cdot}, \ldots, c_{k,\cdot}$, and $c_{\cdot,1}, \ldots, c_{\cdot,k}$. Given this assumption, the $p$-th raw moment is $\mathbb{E}\, c_{i,j}^p = c_{i,\cdot}^p\, c_{\cdot,j}^p / n^p$. We can thus introduce the following property.

**Definition 5** (Property [E0]) For all $s_1, \ldots, s_k > 0$, $t_1, \ldots, t_k > 0$, and the random variable $\mathbf{C} \in \mathfrak{C}^{k\times k}$ generated from the hypergeometric model having $c_{i,\cdot} = s_i$ and $c_{\cdot,j} = t_j$ for all $i, j$, we have $\mathbb{E}\, \mathrm{I}(\mathbf{C}) = 0$.

Given any index I, its adjusted-for chance version can be constructed by taking:

$$\mathrm{AI}(\mathbf{C}) = \frac{\mathrm{I}(\mathbf{C}) - \widetilde{\mathrm{I}}(\mathbf{C})}{\overline{\mathrm{I}}(\mathbf{C}) - \widetilde{\mathrm{I}}(\mathbf{C})}, \tag{18}$$

where $\overline{\mathrm{I}}$ is the maximal index value (e.g., 1 for indices fulfilling [B1]) and $\widetilde{\mathrm{I}}$ is the expected index given the assumed randomness model. This way, the maximal value of AI is 1 and its expectation is 0 when two partitions are unrelated.

**Example 3** Based on the relation:

$$\mathbb{E}\left(\sum_{i=1}^{k}\sum_{j=1}^{k}\binom{c_{i,j}}{2}\right) = \frac{\left(\sum_{i=1}^{k}\binom{c_{i,\cdot}}{2}\right)\left(\sum_{j=1}^{k}\binom{c_{\cdot,j}}{2}\right)}{\binom{n}{2}}, \tag{19}$$

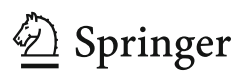

denoting the expected Rand index with $\widetilde{\mathrm{R}}(\mathbf{C})$, we get the *adjusted Rand index* proposed by Hubert and Arabie (1985):

$$\begin{aligned}\mathrm{AR}(\mathbf{C}) &= \frac{\mathrm{R}(\mathbf{C}) - \widetilde{\mathrm{R}}(\mathbf{C})}{1 - \widetilde{\mathrm{R}}(\mathbf{C})} \\ &= \frac{\binom{n}{2}\sum_{i=1}^{k}\sum_{j=1}^{k}\binom{c_{i,j}}{2} - \sum_{i=1}^{k}\binom{c_{i,\cdot}}{2}\sum_{j=1}^{k}\binom{c_{\cdot,j}}{2}}{\binom{n}{2}\frac{1}{2}\left(\sum_{i=1}^{k}\binom{c_{i,\cdot}}{2} + \sum_{j=1}^{k}\binom{c_{\cdot,j}}{2}\right) - \sum_{i=1}^{k}\binom{c_{i,\cdot}}{2}\sum_{j=1}^{k}\binom{c_{\cdot,j}}{2}} \\ &= 1 - \frac{\sum_{i=1}^{k} c_{i,\cdot}^2 + \sum_{j=1}^{k} c_{\cdot,j}^2 - 2\sum_{i=1}^{k}\sum_{j=1}^{k} c_{i,j}^2}{\frac{n+1}{n-1}\left(\sum_{i=1}^{k} c_{i,\cdot}^2 + \sum_{j=1}^{k} c_{\cdot,j}^2\right) - \frac{2}{n-1}\left(n^2 + \frac{1}{n}\sum_{i=1}^{k} c_{i,\cdot}^2 \sum_{j=1}^{k} c_{\cdot,j}^2\right)}.\end{aligned} \tag{20}$$

The FM index can be adjusted in a similar manner, leading to:

$$\begin{aligned}\mathrm{AFM}(\mathbf{C}) &= \frac{\mathrm{FM}(\mathbf{C}) - \widetilde{\mathrm{FM}}(\mathbf{C})}{1 - \widetilde{\mathrm{FM}}(\mathbf{C})} \\ &= \frac{\binom{n}{2}\sum_{i=1}^{k}\sum_{j=1}^{k}\binom{c_{i,j}}{2} - \sum_{i=1}^{k}\binom{c_{i,\cdot}}{2}\sum_{j=1}^{k}\binom{c_{\cdot,j}}{2}}{\binom{n}{2}\sqrt{\sum_{i=1}^{k}\binom{c_{i,\cdot}}{2}\sum_{j=1}^{k}\binom{c_{\cdot,j}}{2}} - \sum_{i=1}^{k}\binom{c_{i,\cdot}}{2}\sum_{j=1}^{k}\binom{c_{\cdot,j}}{2}} \\ &= \frac{\left(\sum_{i=1}^{k}\sum_{j=1}^{k} c_{i,j}^2 - n\right) - \frac{\left(\sum_{i=1}^{k} c_{i,\cdot}^2 - n\right)\left(\sum_{j=1}^{k} c_{\cdot,j}^2 - n\right)}{n(n-1)}}{\sqrt{\left(\sum_{i=1}^{k} c_{i,\cdot}^2 - n\right)\left(\sum_{j=1}^{k} c_{\cdot,j}^2 - n\right)} - \frac{\left(\sum_{i=1}^{k} c_{i,\cdot}^2 - n\right)\left(\sum_{j=1}^{k} c_{\cdot,j}^2 - n\right)}{n(n-1)}}.\end{aligned} \tag{21}$$

Note that Eqs. 20 and 21 are very similar: the difference is that in the former, we have the arithmetic mean in the denominator, whilst in the latter, we see the geometric mean of two terms. In practice, the two indices behave very similarly (see Sect. 4).

**Example 4** An adjusted version of the NMI score was studied by Vinh et al. (2010) (who denoted it by $\mathrm{AMI}_{\mathrm{sum}}$). By noting that:

$$\begin{aligned}\widetilde{\mathrm{MI}}(\mathbf{C}) &= \mathbb{E}\left(\sum_{i=1}^{k}\sum_{j=1}^{k}\frac{c_{i,j}}{n}\log\frac{n\,c_{i,j}}{c_{i,\cdot}\,c_{\cdot,j}}\right) = \\ &= \sum_{i=1}^{k}\sum_{j=1}^{k}\sum_{\ell=\max\{0,\,c_{i,\cdot}+c_{\cdot,j}-n\}}^{\min\{c_{i,\cdot},\,c_{\cdot,j}\}}\frac{\ell}{n}\,\frac{\binom{c_{i,\cdot}}{\ell}\binom{c_{\cdot,j}}{\ell}}{\binom{n}{c_{i,\cdot}}\binom{n}{c_{\cdot,j}}}\,\frac{n!\,\ell!}{(n-c_{i,\cdot}-c_{\cdot,j}+\ell)!}\,\log\frac{n\,\ell}{c_{i,\cdot}\,c_{\cdot,j}},\end{aligned} \tag{22}$$

we obtain:

$$\begin{aligned}\mathrm{AMI}(\mathbf{C}) &= \frac{\mathrm{NMI}(\mathbf{C}) - \widetilde{\mathrm{NMI}}(\mathbf{C})}{1 - \widetilde{\mathrm{NMI}}(\mathbf{C})} \\ &= \frac{\sum_{i=1}^{k}\sum_{j=1}^{k}\frac{c_{i,j}}{n}\log\frac{n\,c_{i,j}}{c_{i,\cdot}\,c_{\cdot,j}} - \widetilde{\mathrm{MI}}(\mathbf{C})}{\frac{1}{2}\left(\sum_{i=1}^{k}\frac{c_{i,\cdot}}{n}\log\frac{n}{c_{i,\cdot}} + \sum_{j=1}^{k}\frac{c_{\cdot,j}}{n}\log\frac{n}{c_{\cdot,j}}\right) - \widetilde{\mathrm{MI}}(\mathbf{C})},\end{aligned} \tag{23}$$

being a formula we will not be expanding further because of its complexity.

None of the three aforementioned adjusted indices are scale-invariant. By construction, they are defined only for integer confusion matrices.

### 3.6 Lower Bounds and Normalisation of Indices

An index that is adjusted for chance has the expected value of 0 for partitions generated at random from an assumed model of randomness. In order for that to be possible, it must take negative values in cases “worse than average when picked at random”.

**Example 5** Canvass the two following matrices with the same corresponding row sums:

$$\mathbf{C} = \begin{bmatrix} 16 & 15 & 11 \\ 9 & 14 & 7 \\ 11 & 10 & 15 \end{bmatrix} \quad \text{and} \quad \mathbf{U} = \begin{bmatrix} 14 & 14 & 14 \\ 10 & 10 & 10 \\ 12 & 12 & 12 \end{bmatrix}.$$

From the current paper’s perspective, the former case is not as undesirable as the latter, where the predicted cluster memberships in each true cluster are assigned uniformly. The adjusted Rand index indicates this correctly: it yields $\mathrm{AR}(\mathbf{C}) = 0$ and $\mathrm{AR}(\mathbf{U}) \approx -0.019$. However, we argue that a negative index value might be difficult to interpret,[3] especially if its lower bound depends on the scale of the confusion matrix.

Instead of looking from a statistical viewpoint, we can take an algebraic perspective, where bounding the index from below, e.g., by 0 could be more informative[4] (Charon et al., 2006; Chacón & Rastrojo, 2023). This is beneficial in the case where we run a clustering algorithm on many benchmark datasets and thus obtain numerous similarity scores that should be aggregated into a single number (e.g., by considering sample quantiles or the arithmetic mean). In such a way, we bring all the indices to the same range, e.g., [0, 1], which is dependent on neither $k$ nor $n$. If the minimum is difficult to obtain or is uninformative, the value of 0 should be attained by confusion matrices that we identify as undesirable outcomes.

We proclaim that there are two worst-case outcomes of a clustering algorithm when its results are compared with a fixed reference partition. The first scenario is where the elements in each row of a confusion matrix are equal to each other:

$$\mathbf{U}^{k\times k}_{s_1,\ldots,s_k} = \begin{bmatrix} s_1/k & s_1/k & \ldots & s_1/k \\ s_2/k & s_2/k & \ldots & s_2/k \\ \vdots & \vdots & \ddots & \vdots \\ s_k/k & s_k/k & \ldots & s_k/k \end{bmatrix}. \tag{24}$$

It corresponds to a clustering method that assigns the cluster memberships uniformly, in an uninformed manner[5].

[3] One of the reviewers disagreed with this statement: “I actually find it very desirable that the latter value tells me that this is worse than average.” We acknowledge that in certain applications, it might indeed be more welcome a behaviour.

[4] Interestingly, Chacón (2021) suggested that if a measure is not normalised, the lower bound should be provided alongside the index value. This way, we can report, e.g., “A=0.53 (min=0.5 for $k = 2$)”, indicating that the similarity score is close to the worst-case scenario.

[5] We note that this matrix is equal to the expected value of the confusion matrix in the hypergeometric model assuming given row sums $s_1, \ldots, s_k > 0$ and each column sum equal to $n/k$. However, in our setting, the sense of a uniform assignment of the predicted cluster memberships is purely algebraic; we do not generate the matrices at random. Similarly, Chacón (2021) notes that this corresponds to “the situation where the labels of the first clustering are perfectly independent of the labels in the second clustering”.

The second undesirable case is where all columns but one are equal to 0:

$$\mathbf{O}^{k\times k}_{s_1,\dots,s_k} = \begin{bmatrix} s_1 & 0 & \dots & 0 \\ s_2 & 0 & \dots & 0 \\ \vdots & \vdots & \ddots & \vdots \\ s_k & 0 & \dots & 0 \end{bmatrix}. \tag{25}$$

It represents predicted partitions where all the points are allocated to the same (one[6]) cluster, assuming given true cluster sizes $s_1, \dots, s_k > 0$.

These two cases lead us to the following desirable properties.

**Definition 6** (Property [U0]) For all $s_1, \dots, s_k > 0$, we have $\mathrm{I}(\mathbf{U}^{k\times k}_{s_1,\dots,s_k}) = 0$.

**Definition 7** (Property [O0]) For all $s_1, \dots, s_k > 0$, we have $\mathrm{I}(\mathbf{O}^{k\times k}_{s_1,\dots,s_k}) = 0$.

In Table 1, we have summarised the index values for matrices given by Eqs. 24 and 25. The derivations are quite elementary; hence, they were omitted. Based on these results, we can imply that out of the indices considered so far, only MI and NMI fulfil [U0]. On the other hand, [O0] is true for MI, NMI, AR, AFM, and AMI.

From Table 1, we see that for matrices like $\mathbf{U}$ (24), AR, AFM, and AMI actually take negative values; Chacón and Rastrojo (2023) give the formula for the minimum of AR. Furthermore, R′, FM′, and A are bounded from below by $1/k$; see Appendix A.2 for a proof in the case of the A index. Thus, that the lowest possible value of an index is equal to 0 must be introduced as a separate property.

**Definition 8** (Property [B0]) For all $s_1, \dots, s_k > 0$, we have that $\min_{\mathbf{C}\in\mathfrak{C}^{k\times k}_{s_1,\dots,s_k}} \mathrm{I}(\mathbf{C}) = 0$.

Out of the indices considered so far, only MI and NMI fulfil [B0]. Note that even if an index fulfils both [U0] and [O0], there is no guarantee that it is bounded from below by 0.

**Example 6** Consider the limiting version of AR (which is a formula equivalent to the one proposed by Morey and Agresti (1984)),

$$\begin{aligned} \mathrm{NR}'(\mathbf{C}) &= \lim_{s\to\infty} \mathrm{AR}(s\mathbf{C}) \\ &= \frac{\sum_{i=1}^k \sum_{j=1}^k c_{i,j}^2 - \frac{\sum_{i=1}^k c_{i,\cdot}^2 \sum_{j=1}^k c_{\cdot,j}^2}{n^2}}{\frac{\sum_{i=1}^k c_{i,\cdot}^2 + \sum_{j=1}^k c_{\cdot,j}^2}{2} - \frac{\sum_{i=1}^k c_{i,\cdot}^2 \sum_{j=1}^k c_{\cdot,j}^2}{n^2}}, \end{aligned} \tag{26}$$

as well as the limiting version of AFM:

$$\begin{aligned} \mathrm{NFM}'(\mathbf{C}) &= \lim_{s\to\infty} \mathrm{AFM}(s\mathbf{C}) \\ &= \frac{\sum_{i=1}^k \sum_{j=1}^k c_{i,j}^2 - \frac{\sum_{i=1}^k c_{i,\cdot}^2 \sum_{j=1}^k c_{\cdot,j}^2}{n^2}}{\sqrt{\sum_{i=1}^k c_{i,\cdot}^2 \sum_{j=1}^k c_{\cdot,j}^2} - \frac{\sum_{i=1}^k c_{i,\cdot}^2 \sum_{j=1}^k c_{\cdot,j}^2}{n^2}}. \end{aligned} \tag{27}$$

[6] If we had an index that does not accept matrices with column sums of 0, we would introduce $\mathbf{O} = (o_{i,j})$ as having $o_{i,i} = \varepsilon > 0$ and $o_{i,1} = s_i - \varepsilon$ for all $i \geq 2$ and some infinitesimal $\varepsilon > 0$. However, all our examples are well-defined for $\mathbf{O}$s given by Eq. 25, which is, by all odds, a much simpler setting.

**Table 1** Values of indices for confusion matrices given by Eqs. 24 (uniform assignment; see property [U0]) and 25 (all points assigned to a single cluster; see [O0]), where $S^2 = \sum_{i=1}^{k} s_i^2$. The R, FM, AR, AFM, and AMI indices assume that input matrices are integer

| Index | $\mathbf{U}^{k\times k}_{s_1,\ldots,s_k}$ | $\mathbf{U}^{k\times k}_{s,\ldots,s}$ | $\mathbf{O}^{k\times k}_{s_1,\ldots,s_k}$ | $\mathbf{O}^{k\times k}_{s,\ldots,s}$ |
|---|---|---|---|---|
| R (2) | $1 - \frac{n^2+(k-2)S^2}{kn(n-1)}$ | $1 - \frac{2(k-1)}{k^2(1-1/n)}$ | $\frac{S^2-n}{n(n-1)}$ | $\frac{n/k-1}{n-1}$ |
| R′ (15) | $1 - \frac{n^2+(k-2)S^2}{kn^2}$ | $1 - \frac{2(k-1)}{k^2}$ | $\frac{S^2}{n^2}$ | $\frac{1}{k}$ |
| FM (4) | $\frac{S^2-kn}{\sqrt{kn(n-k)(S^2-n)}}$ | $\frac{1/k^2-1/n}{1/k-1/n}$ | $\sqrt{\frac{S^2-n}{n(n-1)}}$ | $\sqrt{\frac{1/k-1/n}{1-1/n}}$ |
| FM′ (16) | $\sqrt{\frac{S^2}{kn^2}}$ | $\frac{1}{k}$ | $\sqrt{\frac{S^2}{n^2}}$ | $\sqrt{\frac{1}{k}}$ |
| MI (6) | 0 | 0 | 0 | 0 |
| A (12) | $\frac{1}{k}$ | $\frac{1}{k}$ | $\frac{\max_i s_i}{n}$ | $\frac{1}{k}$ |
| BA (36) | $\frac{1}{k}\sum_{i=1}^{k}\frac{s_i}{\max\{ks_i,n\}}$ | $\frac{1}{k}$ | $\frac{\max_i s_i}{kn}$ | $\frac{1}{k^2}$ |
| CA (34) | $\frac{1}{k}$ | $\frac{1}{k}$ | $\frac{1}{k}$ | $\frac{1}{k}$ |
| AR (20) | $< 0$ | $\frac{2-k-1/k}{n(1-1/k)-(k-1)} < 0$ | 0 | 0 |
| AFM (21) | $< 0$ | $\frac{2-k-1/k}{n(1-1/k)-(k-1)} < 0$ | 0 | 0 |
| AMI (23) | $< 0$ | $< 0$ | 0 | 0 |
| NR′ (26) | 0 | 0 | 0 | 0 |
| NFM′ (27) | 0 | 0 | 0 | 0 |
| NMI (17) | 0 | 0 | 0 | 0 |
| NA (30) | 0 | 0 | $\frac{k\max_i s_i-n}{n(k-1)}$ | 0 |
| NCR′ (31) | 0 | 0 | 0 | 0 |
| NCFM′ (32) | 0 | 0 | 0 | 0 |
| NCMI (33) | 0 | 0 | 0 | 0 |
| NBA (37) | 0 | 0 | 0 | 0 |
| NCA (35) | 0 | 0 | 0 | 0 |

The exact formulae denoted by standalone "$< 0$"s were omitted due to their complexity (they are all negative and approach 0 as $n \to \infty$)

At the cost of [E0], we have gained [SU], [U0], and [O0] (compare Table 1). However, [B0] does not hold; for instance, if:

$$\Omega = \begin{bmatrix} 50 & 25 \\ 25 & 0 \end{bmatrix}, \tag{28}$$

we have $\mathrm{NR}'(\Omega) = \mathrm{NFM}'(\Omega) = -\frac{1}{15}$.

If we are able to identify the minimum of an index I over all matrices with given row sums, denoted by $\underline{\mathrm{I}}$, we can introduce its normalised version by taking:

$$\mathrm{NI}(\mathbf{C}) = \frac{\mathrm{I}(\mathbf{C}) - \underline{\mathrm{I}}(\mathbf{C})}{\overline{\mathrm{I}}(\mathbf{C}) - \underline{\mathrm{I}}(\mathbf{C})}, \tag{29}$$

where again $\overline{\mathrm{I}}$ is the maximal index value. This way we obtain the index that has the minimum value of 0 and maximum of 1.

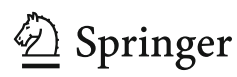

**Example 7** For all $s_1, \ldots, s_k > 0$, we have $\min_{\mathbf{C}\in\mathfrak{C}^{k\times k}_{s_1,\ldots,s_k}} \mathrm{A}(\mathbf{C}) = 1/k$; see Appendix A.2.[7] We can thus introduce the normalised variant of the pivoted accuracy:

$$\mathrm{NA}(\mathbf{C}) = \max_{\sigma\in\mathfrak{S}_k} \frac{\sum_{i=1}^{k} \frac{c_{i,\sigma(i)}}{n} - \frac{1}{k}}{1-\frac{1}{k}} = \max_{\sigma\in\mathfrak{S}_k} \frac{1}{k-1}\left(\sum_{i=1}^{k} \frac{c_{i,\sigma(i)}}{n/k} - 1\right). \tag{30}$$

NA is an example of an index that fulfils [U0] and [B0], but not [O0] (compare Table 1).

### 3.7 Invariance to Cluster Sizes

Here is a more restrictive version of scale invariance: we posit that increasing the number of points in any reference cluster and assigning the new points to the predicted sets without disturbing the proportions of allocations should not change a given cluster validity index; compare the paper by Rezaei and Fränti (2016), who studied a similar postulate. For instance, if 50% of the points are correctly classified in the first reference cluster, then it should not matter whether its cardinality is, say, $c_{1,\cdot} = 100$ or $c_{1,\cdot} = 10{,}000$.

**Definition 9** (Property [SC]) For all $\mathbf{C} \in \mathfrak{C}^{k\times k}$ and every $s_1, \ldots, s_k > 0$, we have $\mathrm{I}(\mathbf{SC}) = \mathrm{I}(\mathbf{C})$, where $\mathbf{S} = \mathrm{diag}(s_1, \ldots, s_k)$.

Note that $\mathbf{SC}$ is a version of the original matrix whose $i$-th row is multiplied by $s_i$.

If an index satisfies [SC], we will say that it is invariant to inequalities in the cluster sizes. Clearly, [SC] implies [SU]. Unfortunately, none of the aforementioned indices fulfils [SC]. However, if an index I satisfies [SU], it is natural to consider its corrected-for-cluster-sizes version CI given for any $\mathbf{C} \in \mathfrak{C}^{k\times k}$ by $\mathrm{CI}(\mathbf{C}) = \mathrm{I}(\mathbf{SC})$, where $\mathbf{S} = \mathrm{diag}(1/c_{1,\cdot}, \ldots, 1/c_{k,\cdot})$.

**Example 8** We can introduce the following indices:

$$\mathrm{NCR}'(\mathbf{C}) = \mathrm{NR}'(\mathbf{SC}), \tag{31}$$

$$\mathrm{NCFM}'(\mathbf{C}) = \mathrm{NFM}'(\mathbf{SC}), \tag{32}$$

$$\mathrm{NCMI}(\mathbf{C}) = \mathrm{NMI}(\mathbf{SC}), \tag{33}$$

where $\mathbf{S} = \mathrm{diag}(1/c_{1,\cdot}, \ldots, 1/c_{k,\cdot})$.

They have all gained [SC] at the cost of losing [SYM]. Interestingly, compared with their original counterparts, NCR′ and NCFM′ now additionally fulfil the previously missing [B0] (see Appendix A.1 for the proof).

**Example 9** Using the same transformation, but this time applied on the pivoted accuracy given by Eq. 12, we can also introduce the *clustering accuracy*:

$$\mathrm{CA}(\mathbf{C}) = \max_{\sigma\in\mathfrak{S}_k} \frac{1}{k} \sum_{i=1}^{k} \frac{c_{i,\sigma(i)}}{c_{i,\cdot}}. \tag{34}$$

It is the average of the proportions of correctly classified points in each true cluster.

[7] Note again that, for simplicity, we consider real-valued confusion matrices. The minimum in the case of crisp memberships was derived by Charon et al. (2006). It is equal to $(\lceil n/k \rceil)/n$ and it is approximately equal to $1/k$ for large $n$.

**Example 10** As $\min_{\mathbf{C}\in\mathcal{C}^{k\times k}_{s_1,\dots,s_k}} \mathrm{CA}(\mathbf{C}) = 1/k$ (see Appendix A.2), we can finally arrive at the new *normalised clustering accuracy* that we announced in the introduction, which is the normalised version of CA:

$$\mathrm{NCA}(\mathbf{C}) = \max_{\sigma\in\mathfrak{S}_k} \frac{\frac{1}{k}\sum_{i=1}^{k}\frac{c_{i,\sigma(i)}}{c_{i,\cdot}} - \frac{1}{k}}{1-\frac{1}{k}} = \max_{\sigma\in\mathfrak{S}_k} \frac{1}{k-1}\left(\sum_{i=1}^{k}\frac{c_{i,\sigma(i)}}{c_{i,\cdot}} - 1\right). \tag{35}$$

NCA fulfils [PER], [SU], [SC], [B1], [U0], [O0], and [B0].

**Remark 4** On a side note, a different type of scaling was considered by Rezaei and Fränti (2016, Sec. 6.1). It is based on the Braun-Blanuqet similarity coefficient (Braun-Blanquet, 1932) and leads to the index given by:

$$\mathrm{BA}(\mathbf{C}) = \max_{\sigma\in\mathfrak{S}_k} \frac{1}{k}\sum_{i=1}^{k}\frac{c_{i,\sigma(i)}}{\max\{c_{i,\cdot}, c_{\cdot,\sigma(i)}\}}. \tag{36}$$

It does not guarantee the fulfilment of our [SC], but preserves symmetry. The authors suggested a normalisation via:

$$\widetilde{\mathrm{BA}}(\mathbf{C}) = \frac{1}{k}\sum_{i=1}^{k}\frac{c_{(i),\cdot}\ c_{\cdot,(i)}}{n\ \max\{c_{(i),\cdot}, c_{\cdot,(i)}\}},$$

where $c_{(i),\cdot}$ is the $i$-th largest row sum and $c_{\cdot,(i)}$ is the $i$-th largest column sum, which leads to the index:

$$\mathrm{NBA}(\mathbf{C}) = \frac{\mathrm{BA}(\mathbf{C}) - \widetilde{\mathrm{BA}}(\mathbf{C})}{1-\widetilde{\mathrm{BA}}(\mathbf{C})}. \tag{37}$$

It fulfils [U0] and [O0], but fails to meet [B0]; e.g., for the matrix given by Eq. 28, we get $\mathrm{NBA}(\Omega) = -1/3$. Interestingly, the authors overcome this limitation by simple clipping, defining the *pair sets index* as:

$$\mathrm{PSI}(\mathbf{C}) = \max\{0, \mathrm{NBA}(\mathbf{C})\}. \tag{38}$$

### 3.8 Monotonicity

It is not rare in the literature to seek functions like $f : Z \to \mathbb{R}$, which preserve a certain partial order $\preceq$ on a set $Z$, i.e., expect that $f(z) \le f(z')$ whenever $z \preceq z'$. For instance, generalised means (e.g., the arithmetic mean and the median) are nondecreasing in each variable (Bullen, 2003, p. xxvi; Grabisch et al., 2009, Chap. 4), and economic inequality indices (e.g., the Gini or Bonferroni ones) preserve the majorisation relation (are Schur convex, satisfy the principle of progressive transfers; Arnold, 2015, Chap. 4).

In the literature, the monotonicity property of clustering similarity indices is often studied in the context of merging or splitting clusters; see the overview by Arinik et al. (2021). Inspired by an empirical sensitivity analysis by Rezaei and Fränti (2016, Sec. 7.3), we would now like to propose one of the possible ways to formalise the property descriptively formulated as "the more similar the predicted partition is to the reference one, the higher the score should be".

Let us define a class of confusion matrices where the maximal elements in each row lie on the main diagonal.

**Definition 10** We call a matrix $\mathbf{C} \in \mathfrak{C}^{k\times k}$ *diagonally max-dominant*, whenever for all $i$, $j$, we have $c_{i,i} \geq c_{i,j}$.

In our context, such matrices represent the case where there is no doubt as to which predicted cluster corresponds to which reference one. We proclaim that, then, correcting the belongingness of a misclassified point (moving it from cluster $j$ to cluster $i$, when it indeed belongs to $i$), should, at the very least, not result in a decrease of an external cluster validity measure.

Let the partial order $\preceq_{\text{DMD}}$ on $\mathbf{C} \in \mathfrak{C}^{k,k}$ be defined in such a way that $\mathbf{C} \preceq_{\text{DMD}} \mathbf{C}'$ if and only if $\mathbf{C}$ and $\mathbf{C}'$ have identical row sums ($c_{i,\cdot} = c'_{i,\cdot}$, $i = 1, \ldots, k$), are diagonally max-dominant, and $c_{i,i} \leq c'_{i,i}$ for all $i$.

**Definition 11** (Property [MON]) If $\mathbf{C} \preceq_{\text{DMD}} \mathbf{C}'$, then $\mathrm{I}(\mathbf{C}) \leq \mathrm{I}(\mathbf{C}')$.

When I satisfies [MON], then for all diagonally max-dominant $\mathbf{C} \in \mathfrak{C}^{k,k}$, every $i \neq j$, and $c_{i,j} \geq t > 0$, we have $\mathrm{I}(\mathbf{C}) \leq \mathrm{I}(\mathbf{C}')$, where $\mathbf{C}'$ is a version of $\mathbf{C}$ except $c'_{i,i} = c_{i,i} + t$ and $c'_{i,j} = c_{i,j} - t$.

We can define strict monotonicity similarly, by replacing $\leq$ and $\preceq$ with $<$ and $\prec$(i.e., $\preceq$ and not $=$), respectively, in the above definition.

For a diagonally max-dominant matrix $\mathbf{C}$, we have $\max_{\sigma\in\mathfrak{S}_k} \sum_{i=1}^{k} c_{i,\sigma(i)} = \sum_{i=1}^{k} c_{i,i}$ and $\max_{\sigma\in\mathfrak{S}_k} \sum_{i=1}^{k} c_{i,\sigma(i)}/c_{i,\cdot} = \sum_{i=1}^{k} c_{i,i}/c_{i,\cdot}$. Therefore, it is easily seen that A, NA, CA, and NCA are naturally strictly monotonic. Moreover, improving the memberships in the $i$-th true cluster always results in the same change of the index: they increase linearly (the increment depends on $c_{i,\cdot}$ for CA and NCA).

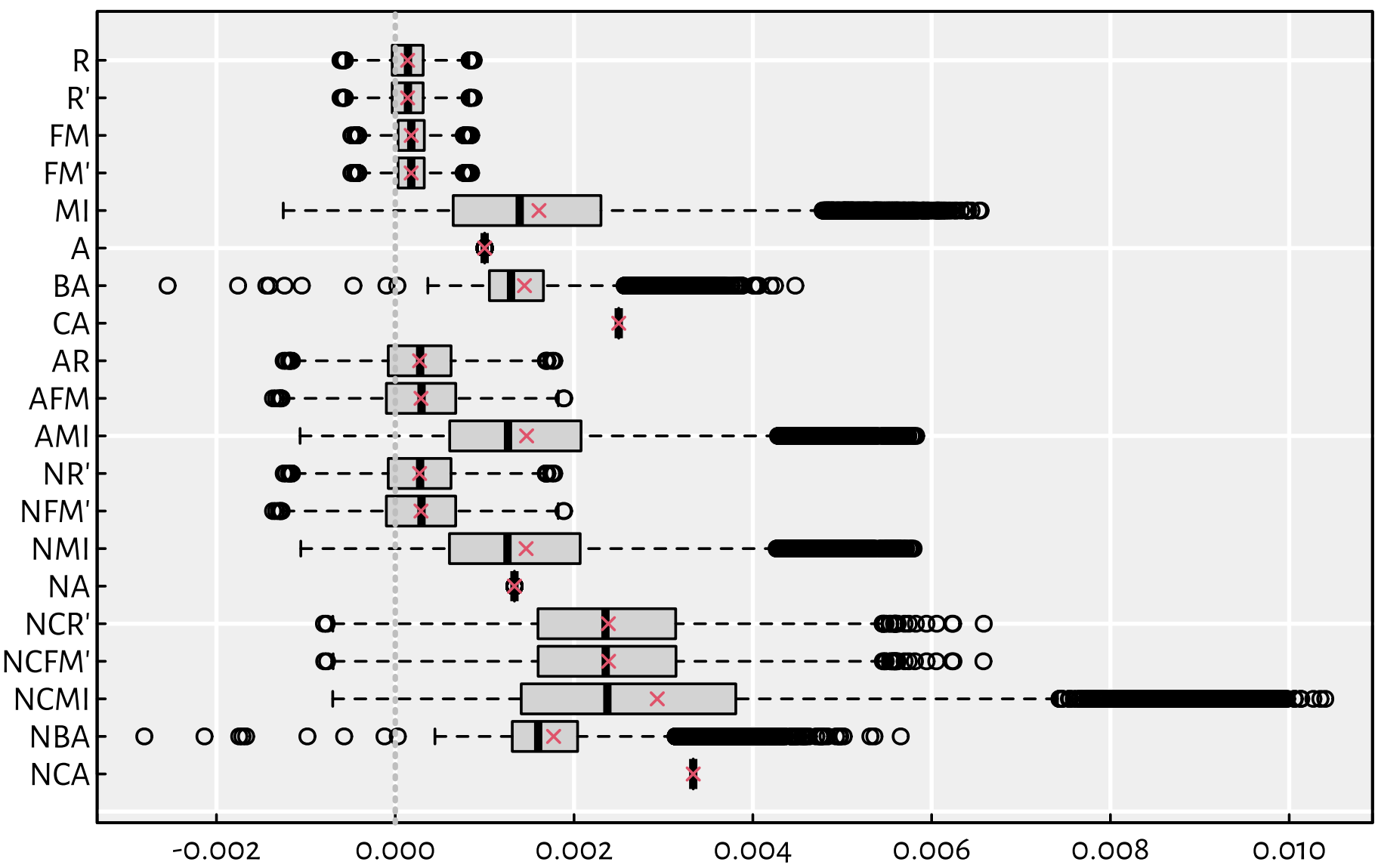

**Fig. 2** Change of index values $\mathrm{I}(\mathbf{C}') - \mathrm{I}(\mathbf{C})$ for randomly generated diagonally max-dominant matrices $\mathbf{C}$ with $k = 4$, $c_{1,\cdot} = c_{2,\cdot} = c_{3,\cdot} = 100$, and $c_{4,\cdot} = 700$, where $\mathbf{C}'$ is the same as $\mathbf{C}$ except that $c'_{1,1} = c_{1,1} + 1$ and $c'_{1,2} = c_{1,2} - 1$ (one point's predicted cluster membership changed). Ideally, an index's response to such an improvement in clustering accuracy should be nonnegative (property [MON]). Amongst the indices studied, only A, CA, NA, and NCA guarantee this Let us now render some counterexamples showing that [MON] does not hold for other indices studied herein.

**Example 11** For any integer $s_1, \ldots, s_k \geq 1$, let $\mathbf{C} \in \mathfrak{C}^{k\times k}$ be a randomly generated matrix whose $i$-th row is obtained using the following procedure:

- Generate $(u_1, \ldots, u_k) \sim \mathrm{Dir}(1, \ldots, 1)$ (a sample from a Dirichlet distribution).
- Let $c_{i,j} = \max\{1, \lfloor u_j s_i \rfloor\}$ for $j = 2, \ldots, k$ and then $c_{i,1} = s_i - (c_{i,2} + \cdots + c_{i,k})$.
- Taking $j = \arg\max_j c_{i,j}$, swap $c_{i,i} \leftrightarrow c_{i,j}$.

This guarantees that $\mathbf{C}$ is diagonally max-dominant, has positive integer elements, and its consecutive row sums are $s_1, \ldots, s_k$.

We generate 10,000 random matrices like $\mathbf{C}$ with $k = 4$ and the row sums $c_{1,\cdot} = c_{2,\cdot} = c_{3,\cdot} = 100$, and $c_{4,\cdot} = 700$, and then improve the membership of only one point, obtaining $\mathbf{C}'$ with $c'_{1,1} = c_{1,1} + 1$ and $c'_{2,1} = c_{2,1} - 1$. Figure 2 presents a box and whisker plot depicting the empirical distribution of the 10,000 differences $\mathrm{I}(\mathbf{C}') - \mathrm{I}(\mathbf{C})$ for each index I. If any increment is negative, and this is the case for all the indices but A, CA, NA, and NCA, then we conclude that [MON] does not hold. Moreover, we note how unpredictably the indices respond to such a simple adjustment of the confusion matrix.

**Table 2** Properties of the indices

| Index | [PER] | [SYM] | [SU] | [SC] | [B1] | [E0] | [U0] | [O0] | [B0] | [MON] |
|---|---|---|---|---|---|---|---|---|---|---|
| R (2) | + | + | − | − | + | − | − | − | − | − |
| R′ (15) | + | + | + | − | + | − | − | − | − | − |
| FM (4) | + | + | − | − | + | − | − | − | − | − |
| FM′ (16) | + | + | + | − | + | − | − | − | − | − |
| MI (6) | + | + | + | − | − | − | + | + | + | − |
| A (12) | + | + | + | − | + | − | − | − | − | + |
| BA (36) | + | + | + | − | + | − | − | − | − | − |
| CA (34) | + | − | + | + | + | − | − | − | − | + |
| AR (20) | + | + | − | − | + | + | − | + | − | − |
| AFM (21) | + | + | − | − | + | + | − | + | − | − |
| AMI (23) | + | + | − | − | + | + | − | + | − | − |
| NR′ (26) | + | + | + | − | + | − | + | + | − | − |
| NFM′ (27) | + | + | + | − | + | − | + | + | − | − |
| NMI (17) | + | + | + | − | + | − | + | + | + | − |
| NA (30) | + | + | + | − | + | − | + | − | + | + |
| NCR′ (31) | + | − | + | + | + | − | + | + | + | − |
| NCFM′ (32) | + | − | + | + | + | − | + | + | + | − |
| NCMI (33) | + | − | + | + | + | − | + | + | + | − |
| NBA (37) | + | + | + | − | + | − | + | + | − | − |
| NCA (35) | + | − | + | + | + | − | + | + | + | + |

R, FM, AR, AFM, AMI assume that input matrices are integer. Our correction for cluster sizes that guarantees [SC] results in the violation of [SYM]. Note that [E0] and [B0] are mutually exclusive

### 3.9 Discussion

Table 2 summarises which index enjoys each of the above properties. Overall, the proposed measure – normalised clustering accuracy (NCA) – fulfils all properties except [SYM] and [E0].

As we have already noted, losing [SYM] is the price we pay for enforcing the invariance to cluster sizes [SC] using the employed normalisation.

Moreover, a nontrivial index cannot have the expected value of 0 for random partitions ([E0]) if it is bounded from below by 0 ([B0]). As far as matrices following the hypergeometric model are concerned, Fig. 3 depicts how the expected values change as a function of $k$ in the case of clusters of equal sizes. Overall, for all indices except R, R′, and MI, we observe a decreasing trend.

**Example 12** To show that most of the properties are actually mild and do not contradict one another, let us consider the index Z given by:

$$\mathrm{Z}(\mathbf{C}) = \begin{cases} 1 \text{ if } \mathbf{C} = \mathbf{P}_\sigma \mathbf{S} \text{ for some } \sigma \in \mathfrak{S}_k \text{ and } \mathbf{S} = \mathrm{diag}(s_1, \ldots, s_k), \\ 0 \text{ otherwise.} \end{cases} \tag{39}$$

Z fulfils all the properties considered except [E0] (however, it is only weakly, not strictly, monotone).

Let us now gain some insight into how to interpret concrete index values.

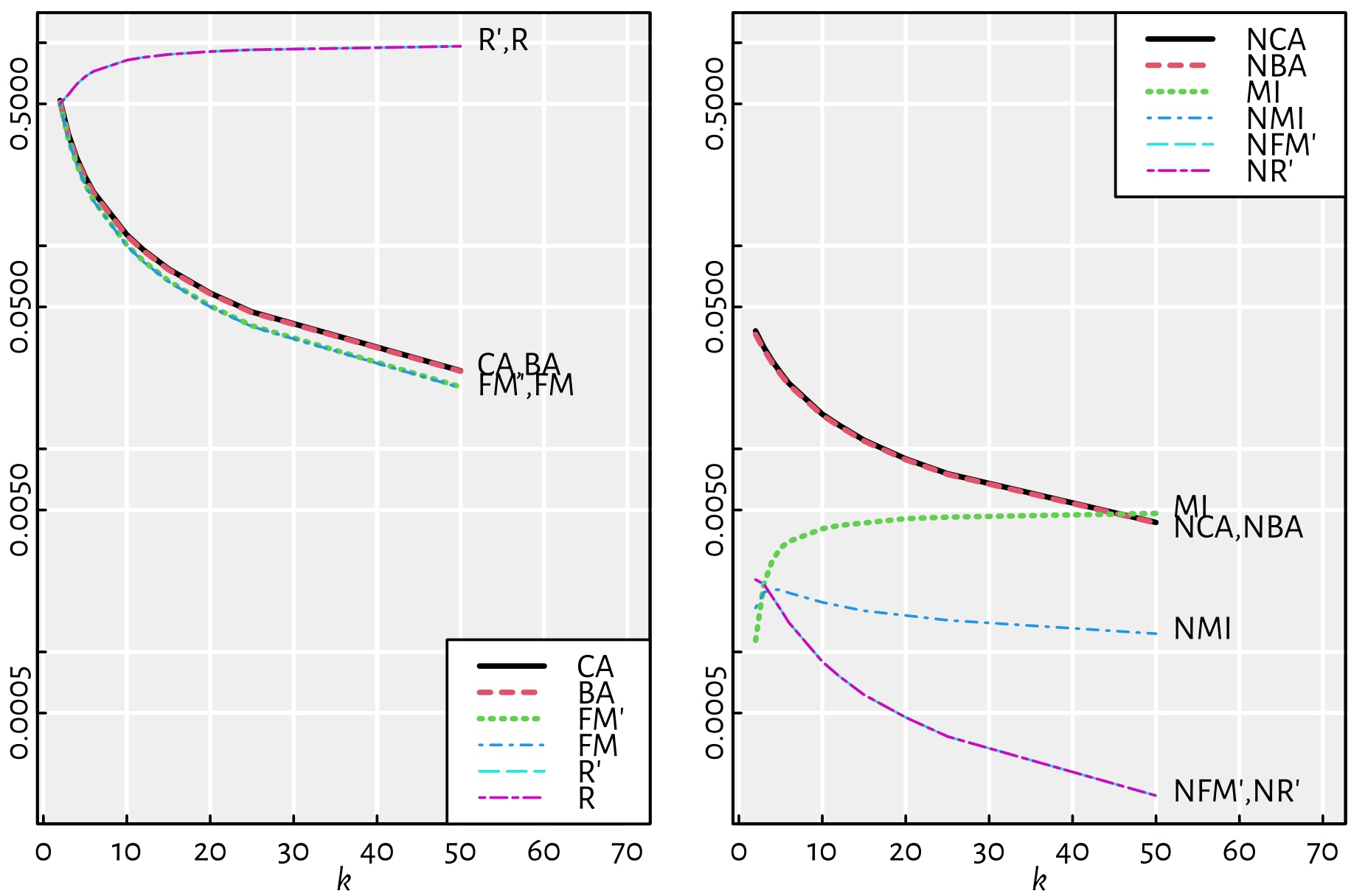

**Fig. 3** Expected values (based on 1,000 Monte Carlo samples) of indices as functions of $k$ for $n = 100k^2$, reference partitions of equal cardinalities, and predicted labels assigned at random (the hypergeometric model). Note the logarithmic scale on the y-axis. AMI, AR, and AFM are adjusted for chance (have expectation 0) and hence were omitted. Recall that, under $c_{1,\cdot} = \cdots = c_{k,\cdot}$, NCFM′=NFM′, NCR′=NR′, NCMI=NMI, NCA=NA, and CA=A. For indices fulfilling [U0] (i.e., those in the right subfigure), we additionally observe that, for all $k$, their expected decrease towards 0

**Example 13** Let $\mathfrak{U}_l^{k\times k}$ represent the class of block diagonal matrices of size $k \times k$ with $l \leq k$ blocks, each consisting of identical values, e.g.:

$$\begin{bmatrix} 100 & 100 & 100 & 0 & 0 & 0 \\ 100 & 100 & 100 & 0 & 0 & 0 \\ 100 & 100 & 100 & 0 & 0 & 0 \\ 0 & 0 & 0 & 300 & 0 & 0 \\ 0 & 0 & 0 & 0 & 150 & 150 \\ 0 & 0 & 0 & 0 & 150 & 150 \end{bmatrix} \in \mathfrak{U}_3^{6\times 6},$$

which represents the identification of $l$ subclusters. In particular, $\mathfrak{U}_k^{k\times k}$ are identity matrices and $\mathfrak{U}_1^{k\times k}$ consist of matrices that we denoted by $\mathbf{U}_{s,\ldots,s}^{k\times k}$ (24). For such matrices, MI, AMI, N(C)MI give different values for each $l$, R′ and R are rather hard to interpret, however:

- A, BA, CA, FM′, and approximately FM yield $l/k$,
- NA, NBA, NCA, N(C)R′, N(C)FM′, and approximately AR and AFM output $(l-1)/(k-1)$.

We have already said that clustering is not classification: clusters are defined up to a permutation of set IDs. The price we pay for normalisation is the loss of 1 "degree of freedom" in the latter formula. This is what makes the normalised measures somewhat less intuitive.

From this perspective, NA and NCA have the most appealing interpretation, because they can be equivalently rewritten as classification rates (averages of proportions of correctly classified points in each cluster) *above* the random cluster membership assignment and *with* the optimal matching of cluster IDs:

$$\mathrm{NA}(\mathbf{C}) = \max_{\sigma \in \mathfrak{S}_k} \frac{1}{k} \sum_{i=1}^{k} \frac{c_{i,\sigma(i)} - \frac{n}{k^2}}{\frac{n}{k} - \frac{n}{k^2}},$$

$$\mathrm{NCA}(\mathbf{C}) = \max_{\sigma \in \mathfrak{S}_k} \frac{1}{k} \sum_{i=1}^{k} \frac{c_{i,\sigma(i)} - \frac{1}{k} c_{i,\cdot}}{c_{i,\cdot} - \frac{1}{k} c_{i,\cdot}},$$

with the difference being that NA relates the current partition to the one where all true clusters are of the same size.

**Example 14** Let us study the behaviour of the normalised indices when we transition between the case of all points allocated to one predicted cluster ($\mathbf{O}_{s_1,\ldots,s_k}^{k\times k}$), through all points correctly classified (identity matrix), to the uniform assignment ($\mathbf{U}_{s_1,\ldots,s_k}^{k\times k}$), one point at a time. Figure 4 depicts the case of $k = 2$ and $s_1 = s_2 = 54$:

1. We start with all the points allocated to cluster 1. All indices fulfil [O0]; therefore, the reported similarities are equal to 0.
2. We move the 54 points from the second reference cluster to its own group. For all indices but NCA, the response is initially slow, and speeds up only at the end.
3. All indices enjoy [B1]; therefore, we reach the similarity of 1.
4. We move 13 (ca. 25%) points back to cluster 1.
5. We spread the points from cluster 1 uniformly.
6. We spread the remaining points from cluster 2 uniformly, arriving at similarity 0 because of [U0].

Figure 5 illustrates $k = 3$ and $s_1 = s_2 = s_3 = 24$:

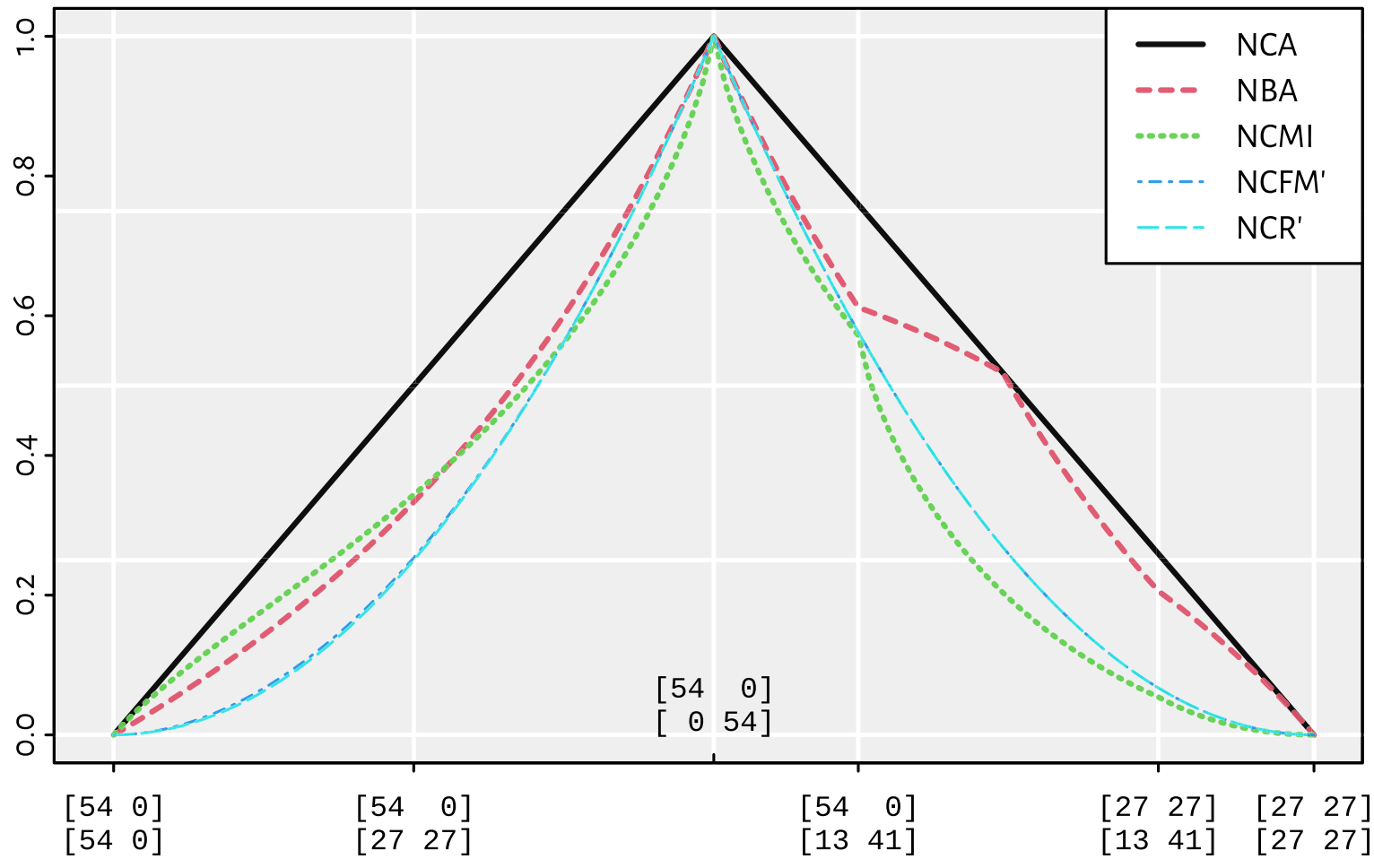

**Fig. 4** Behaviour of selected indices when moving from all points in a single cluster through the perfect match to the uniform assignment, one point at a time ($k = 2$, clusters of equal sizes; hence, NA=NCA, NR′=NCR′, etc.). Only N(C)A yields a predictable, linear response. Note that N(C)R′ and N(C)FM′ are very similar: their curves are almost overlapping

1. We start with all the points allocated to a single cluster.
2. We move all the points in the third true cluster to its own group.
3. We do the same with points from the second group until we reach the perfect match. Notice that NCMI, NCR′, and NCFM′ initially decrease.
4. We move 12 points from the true cluster 1 to cluster 2.
5. We relocate 12 points from the true cluster 2 to cluster 1.
6. We move 8 points from the true cluster 1 to cluster 3 so that they are spread uniformly.
7. We do the same in cluster 2.
8. And similarly in cluster 3.

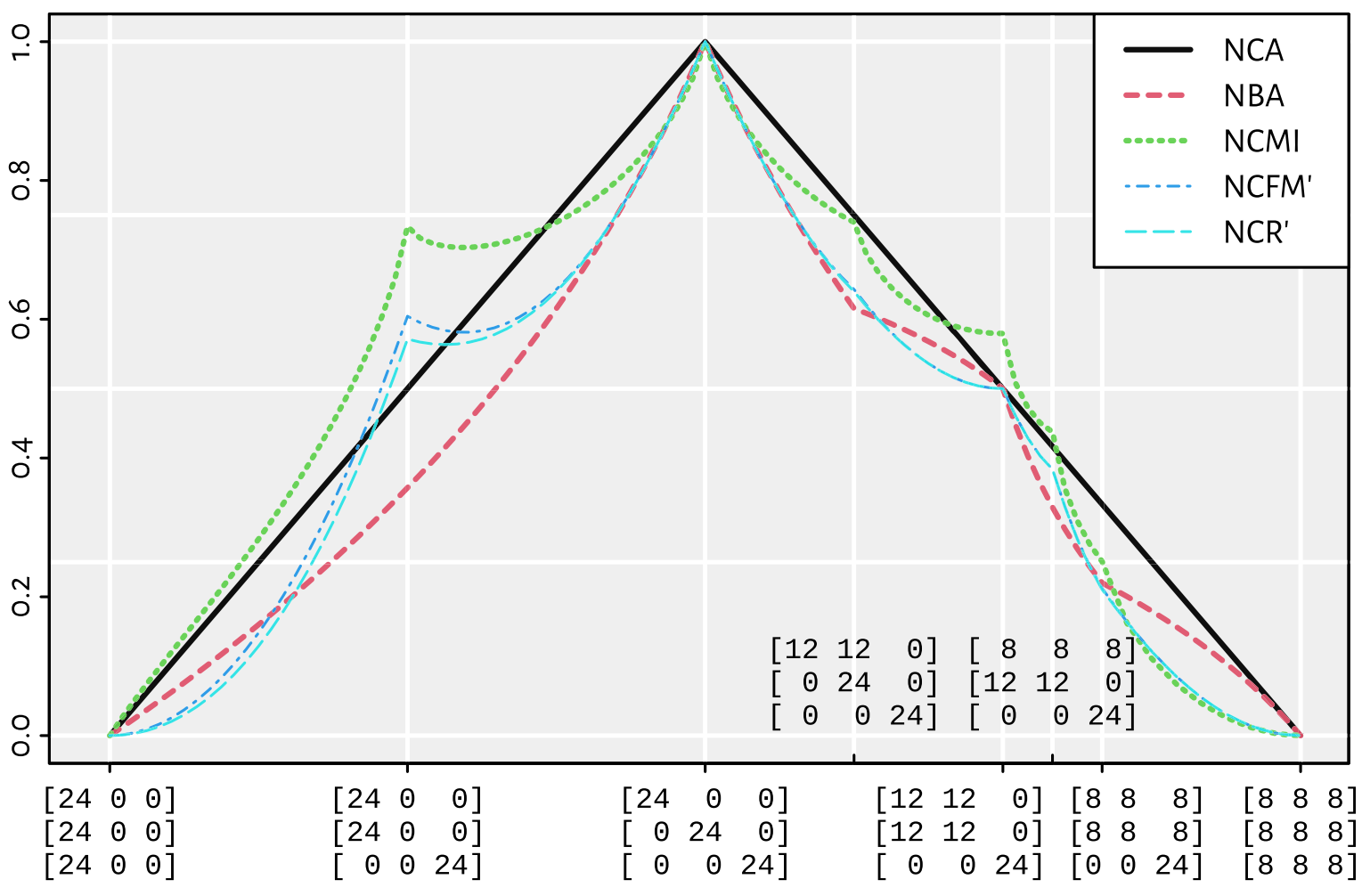

**Fig. 5** Behaviour of selected indices when moving individual points between three clusters ($k = 3$, clusters of equal sizes)

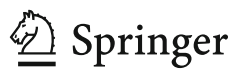

Only A, NA, CA, and NCA change uniformly when improving the cluster memberships.

## 4 Experiments

Let us compare the relationships between the indices on benchmark data. We consider 65 datasets from the paper by Gagolewski (2022) that consist of up to 10,000 points and whose labels do not include any noise points, namely, from the FCPS battery (Thrun & Ultsch, 2020): atom, chainlink, engytime, hepta, lsun, target, tetra, twodiamonds, wingnut; from Graves (Graves & Pedrycz, 2010): dense, fuzzyx, line, parabolic, ring, ring_outliers, zigzag; from SIPU (Fränti & Sieranoja, 2018): a1, a2, a3, aggregation, compound, d31, flame, jain, pathbased, r15, s1, s2, s3, s4, spiral, unbalance; from UCI (Dua & Graff, 2022): ecoli, glass, ionosphere, sonar, statlog, wdbc, wine, yeast; Other: iris, iris5, square; and from WUT: circles, cross, graph, isolation, labirynth, mk1, mk2, mk3, mk4, olympic, smile, stripes, trajectories, trapped_lovers, twosplashes, windows, x1, x2, x3, z1, z2, z3. Each dataset comes with at least one reference partition, which we related, using all the external cluster validity measures studied herein, to the outputs of 10 algorithms: classical agglomerative hierarchical clustering methods based on the Single, Average, Complete, Ward, Centroid, and Median linkages (Müllner, 2011) and algorithms implemented in *scikit-learn* for Python (Pedregosa, 2011): K-Means, expectation-maximisation (EM) for Gaussian mixtures (`n_init=100`), Birch (`threshold=0.01`, `branching_factor=50`), and Spectral (`affinity=Laplacian`, `gamma=5`). In each case, the ground truth labels determine the number of subsets $k$ to be detected, which is given as input to the algorithms. The spectral algorithm failed to converge on one dataset (WUT/x3 with $k = 3$), so we excluded this benchmark scenario from further analysis.

Overall, we obtained $74 \cdot 10 = 740$ readings of each of the similarity scores. We will not take MI into account because it does not meet [B1].

Let us first relate the indices to each other pairwisely (95 cases of perfect agreement between true and predicted labels were removed as this trivially corresponds to all scores equal to 1). There is a very high degree of correlation between specific pairs (Pearson's $r > 0.999$): R and R′; FM and FM′; AFM, NFM′, AR, and NR′; NCFM′ and NCR′; AMI and NMI. Therefore, we will only be considering the last index from each group from now on.

The least correlated pairs of indices, as measured with the Spearman's rank correlation coefficient, which takes into account any monotonic relationships (not necessarily linear) are: FM′ and NCMI ($\rho \approx 0.751$); FM′ and NCA ($\rho \approx 0.766$); BA and NCMI ($\rho \approx 0.766$); FM′ and R′ ($\rho \approx 0.772$); R′ and CA ($\rho \approx 0.777$).

Some non-normalised indices correlate quite strongly with their normalised counterparts. The following index pairs have $\rho > 0.95$: R′ and NR′ ($\rho \approx 0.970$); R′ and NMI ($\rho \approx 0.963$); CA and NCA ($\rho \approx 0.960$); A and NA ($\rho \approx 0.951$). Nevertheless, we note that the true partitions in our benchmark set have diverse cardinalities (ranging from 2 to 50), and the lower bounds of the non-normalised indices are a function of $k$.

Focusing on the normalised indices, the scatter plot matrix in Fig. 6 shows how different indices relate to one another.

Let us note that 29 of the 74 true label vectors define clusters of equal sizes. Overall, for 39 label vectors, the Gini index of true cluster sizes is less than 0.05 (with the average Gini index being 0.189). Therefore, in our study, the correction for cluster sizes has no or very small effect in many cases. This is why NR′ and NCR′, NMI and NCMI, as well as NA and NCA correlate highly with one another.

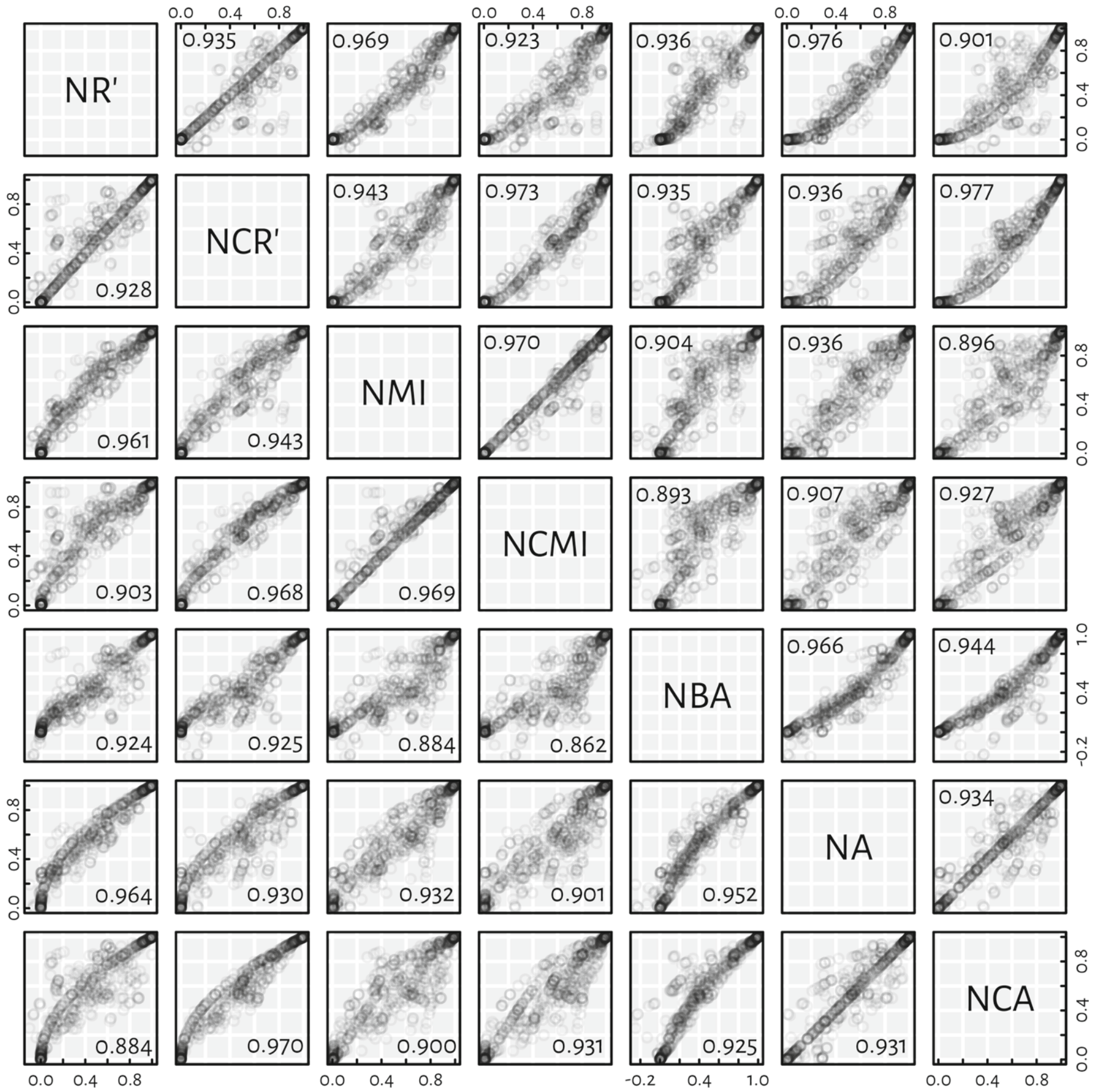

**Fig. 6** Scatter plot matrix for selected normalised indices (we noted that some index groups are very highly correlated: AFM, NFM′, AR, and NR′; NCFM′ and NCR′; AMI and NMI). In the lower right and top left corners, respectively, Pearson's $r$ and Spearman's $\rho$ correlation coefficients are reported. Circa 53% of the true partitions in our data sample consist of clusters of almost equal sizes; hence, the correction for [SC] (e.g., transforming NA to obtain NCA) has little impact in these cases

We note a high degree of rank correlation between: NCR′ and NCA; NCR′ and NCMI; NR′ and NA; NR′ and NMI; NBA and NA. Thence, we can try to express one index from each pair by a monotone function of another with not too big an error.

We may also be interested in determining how the different external validity measures allow us to compare the overall quality of the 10 algorithms. Table 3 gives the rankings based on the median scores. Dasgupta and Ng (2009) and Gagolewski (2022) noted that one dataset can have many possible equally valid clusterings. Thus, in what follows, in the case of datasets with more than one reference labelling, for each algorithm, the best score is taken into account.

**Table 3** Rankings of the 10 clustering algorithms based on median values (in round brackets) of different external cluster validity measures across the 64 benchmark datasets

| | NR′ | NCR′ | NMI | NCMI | NBA | NA | NCA |
|---|---|---|---|---|---|---|---|
| GaussMix | 1 (0.80) | 1 (0.79) | 1 (0.80) | 1 (0.83) | 1 (0.82) | 2 (0.85) | 1 (0.87) |
| Spectral | 2 (0.75) | 2 (0.75) | 2 (0.79) | 2 (0.80) | 2 (0.78) | 1 (0.85) | 2 (0.85) |
| Ward | 4 (0.54) | 4 (0.60) | 7 (0.63) | 7 (0.67) | 5 (0.51) | 5 (0.63) | 3 (0.78) |
| Birch | 3 (0.54) | 3 (0.64) | 4 (0.64) | 5 (0.71) | 4 (0.53) | 4 (0.64) | 4 (0.77) |
| KMeans | 6 (0.51) | 5 (0.58) | 6 (0.64) | 4 (0.72) | 3 (0.54) | 3 (0.68) | 5 (0.72) |
| Average | 5 (0.51) | 6 (0.55) | 5 (0.64) | 6 (0.68) | 6 (0.44) | 6 (0.59) | 6 (0.63) |
| Median | 10 (0.37) | 9 (0.51) | 8 (0.57) | 10 (0.60) | 8 (0.41) | 9 (0.55) | 7 (0.63) |
| Centroid | 7 (0.47) | 7 (0.54) | 9 (0.56) | 8 (0.65) | 7 (0.42) | 7 (0.58) | 8 (0.63) |
| Complete | 9 (0.40) | 8 (0.51) | 10 (0.53) | 9 (0.63) | 9 (0.40) | 8 (0.56) | 9 (0.57) |
| Single | 8 (0.45) | 10 (0.46) | 3 (0.76) | 3 (0.74) | 10 (0.28) | 10 (0.44) | 10 (0.43) |

Let us stress that the purpose of this experiment is not to discover good algorithms, but to assess the effect of index choice on the rankings

Only two methods, Gaussian mixtures and spectral clustering, are consistently ranked as the best ones.

If we employ Kendall's $\tau$, a correlation measure based on counting concordant and discordant object pairs, to assess the similarity of the rankings, we discover that the rankings generated by NMI and NCMI are the least correlated with those by other indices. For instance, they both rank the single linkage method third, whereas most other indices consider it the worst. The most similar pair is NA and NBA. Also, NCR′ correlates highly with NR′, NBA, NA, and NCA.

## 5 Partitions of Different Cardinalities

Many indices can be extended to the case of confusion matrices with the number of rows $k$ being different from the number of columns $k'$. For instance, in the definitions of R, FM, MI, and their derivatives, we can replace the summation $\sum_{i=1}^{k}\sum_{j=1}^{k}$ with $\sum_{i=1}^{k}\sum_{j=1}^{k'}$.

We can generalise the normalised clustering accuracy as:

$$\mathrm{NCA}'(\mathbf{C}) = \max_{\sigma:\{1,\dots,k\}\overset{\text{inject.}}{\to}\{1,\dots,\max\{k,k'\}\}} \frac{1}{k-1}\left(\sum_{i=1}^{k}\frac{c_{i,\sigma(i)}}{c_{i,\cdot}}-1\right), \tag{40}$$

under the assumption $c_{i,j}=0$ for $j>k'$. If $k<k'$, some predicted clusters are not matched with true ones. Hence, they are not counted, which decreases the computed similarity. If $k>k'$, we treat the missing $k-k'$ predicted clusters as empty: the number of points matched is zero. Overall, this version of the index penalises a clustering algorithm that outputs a grouping of cardinality $k'$ different from the desired $k$.

Yet, we might also be interested in the case where $k'\neq k$ is what we actually ask all the benchmarked algorithms for. In such a case, we can consider:

$$\mathrm{NCA}''(\mathbf{C}) = \begin{cases} \max_{\sigma:\{1,\dots,k'\}\overset{\text{onto}}{\to}\{1,\dots,k\}} \frac{1}{k-1}\left(\sum_{j=1}^{k'}\frac{c_{\sigma(j),j}}{c_{\sigma(j),\cdot}}-1\right) & \text{for } k'\geq k,\\ \max_{\sigma:\{1,\dots,k\}\overset{\text{onto}}{\to}\{1,\dots,k'\}} \frac{1}{k-1}\left(\sum_{i=1}^{k}\frac{c_{i,\sigma(i)}}{c_{i,\cdot}}-1\right) & \text{otherwise.} \end{cases}$$

We leave the exploration of the extensions of the indices to the case $k \neq k'$ and their properties as a topic of further research, because it deserves a more exhaustive treatment, for which no room is left in the current contribution.

## 6 Conclusion

We reasoned that using symmetric partition similarity scores to compare predicted partitions with *fixed* reference ones might not be ideal. Thus, we proposed a new measure called normalised clustering accuracy, which uses optimal cluster matching and is corrected for cluster sizes. We showed that it enjoys a number of desirable properties, including scale invariance, boundedness, and monotonicity. As far as interpretability is concerned, e.g., NCA=0.5 means that *above* the perfectly uniform cluster membership assignment, on average, 50% of points in each cluster are correctly discovered.

As a topic for further research, we would like to use some of the proposed properties as a basis for characterising certain indices. We would also like to study other desirable features discussed in the literature (Hennig, 2015; Wagner & Wagner, 2006; Warrens & van der Hoef, 2022; Rezaei & Fränti, 2016; Luna-Romera et al., 2019; Gates et al., 2019; Arinik et al., 2021; Xiang, 2012). It will also be interesting to derive the formula for the expected value of $\max_\sigma \sum_i c_{i,\sigma(i)}$ and $\max_\sigma \sum_i c_{i,\sigma(i)}/c_{i,\cdot}$ in the hypergeometric and other random models.

Furthermore, we will extend our notes from Sect. 5 to the case where the predicted clustering can be finer- or coarser-grained than the reference one (partitions of different cardinalities), including whole cluster hierarchies. We will also formulate similar properties that are more tailored to comparing fuzzy/soft/probabilistic (e.g., overlapping) or other types of partitions (Horta & Campello, 2015; Campagner et al., 2023; Andrews et al., 2022; D'Ambrosio et al., 2021; Hüllermeier et al., 2012), also in the case where the reference clustering is not crisp (compare Flynt et al., 2019).

Moreover, we may want to study similar metrics adjusted to the problem of semi-supervised learning, i.e., where some cluster memberships are known to the algorithm a priori.

## Appendix

### A.1 Proof that NCR′ and NCFM′ Fulfil [B1] and [B0]

As $\sqrt{\sum_{i=1}^k c_{i,\cdot}^2 \sum_{j=1}^k c_{\cdot,j}^2} \leq \frac{\sum_{i=1}^k c_{i,\cdot}^2 + \sum_{j=1}^k c_{\cdot,j}^2}{2}$, we have $\mathrm{NR}'(\mathbf{C}) \leq \mathrm{NFM}'(\mathbf{C}) \leq 1$. The latter inequality is due to the fact that $\sum_{i=1}^k \sum_{j=1}^k c_{i,j}^2 \leq \sqrt{\sum_{i=1}^k c_{i,\cdot}^2 \sum_{j=1}^k c_{\cdot,j}^2}$, which holds because $\sum_{i=1}^k c_{i,\cdot}^2 = \sum_{i=1}^k \sum_{j=1}^k c_{i,j}^2 + \sum_{i=1}^k \sum_{u \neq v} c_{i,u} c_{i,v}$ and $\sum_{j=1}^k c_{j,\cdot}^2 = \sum_{i=1}^k \sum_{j=1}^k c_{i,j}^2 + \sum_{j=1}^k \sum_{u \neq v} c_{u,j} c_{v,j}$. This also easily implies that $\mathrm{NR}'(\mathbf{C}) = \mathrm{NFM}'(\mathbf{C}) = 1$ if and only if $\mathbf{C} = \mathbf{P}_\sigma \mathbf{S}$ for some $\sigma \in \mathfrak{S}_k$ and $\mathbf{S} = \mathrm{diag}(s_1, \ldots, s_k)$. The same holds for NCR′ and NCFM′. Hence, they meet property [B1].

To show that NCFM′ and NCR′ are nonnegative, it suffices to prove that $\frac{\sum_{i=1}^k c_{i,\cdot}^2 \sum_{j=1}^k c_{\cdot,j}^2}{n^2} \leq \sum_{i=1}^k \sum_{j=1}^k c_{i,j}^2$ under $c_{1,\cdot} = \cdots = c_{k,\cdot} = 1$, i.e., for matrices adjusted for cluster sizes. As $n = k$, the inequality can be rewritten as $\sum_{j=1}^k \left( \sum_{i=1}^k c_{i,j} \right)^2 \leq \sum_{j=1}^k \left( k \sum_{i=1}^k c_{i,j}^2 \right)$ and that $\left(\sum_{i=1}^{k} c_{i,j}\right)^2 \leq k \sum_{i=1}^{k} c_{i,j}^2$ holds for all $j$ is implied by the Cauchy-Schwarz inequality which states that $\mathbf{u} \circ \mathbf{v} \leq \|\mathbf{u}\| \, \|\mathbf{v}\|$ with $\mathbf{u} = (c_{1,j}, \ldots, c_{k,j})$ and $\mathbf{v} = (1, \ldots, 1)$. As NCFM′ and NCR′ fulfil [U0] and [O0] (compare Table 1), this implies that they also meet [B0].

### A.2 Proof that A Is Bounded from Below by 1/*k*

We want to prove that for all $\mathbf{C} \in \mathfrak{C}^{k \times k}$, we have $\mathrm{A}(\mathbf{C}) = \max_{\sigma \in \mathfrak{S}_k} \sum_{j=1}^{k} \frac{c_{\sigma(j),j}}{n} \geq 1/k$.

Let $\sigma_1 \in \mathfrak{S}_k$ be the permutation that maximises $\sum_{j=1}^{k} c_{\sigma_1(j),j}$. We have to show that $k \sum_{j=1}^{k} c_{\sigma_1(j),j} \geq \sum_{i=1}^{k} \sum_{j=1}^{k} c_{i,j} = n$. Take any other $k-1$ permutations $\sigma_2, \ldots, \sigma_k \in \mathfrak{S}_k$ such that the set $\{(\sigma_i(j), j)\}_{i,j=1,\ldots,k}$ consists of all possible index pairs, $\{1, \ldots, k\} \times \{1, \ldots, k\}$.

As $\sigma_1$ is the maximal permutation, we have $\sum_{j=1}^{k} c_{\sigma_1(j),j} \geq \sum_{j=1}^{k} c_{\sigma_i(j),j}$ for all $i$. Hence, $k \sum_{j=1}^{k} c_{\sigma_1(j),j} \geq \sum_{i=1}^{k} \sum_{j=1}^{k} c_{\sigma_i(j),j} = \sum_{i=1}^{k} \sum_{j=1}^{k} c_{i,j}$, QED.

Note that this also implies that CA is bounded from below by $1/k$ and that NCA is bounded from below by 0. These are actually minima; as they are attained at $\mathbf{U}^{k \times k}_{s_1,\ldots,s_k}$; see Table 1.

**Acknowledgements** The author is indebted to the Editor and Reviewers for providing numerous insightful remarks that led to the improvement of earlier versions of this paper.

**Funding** This research was partially supported by the Australian Research Council Discovery Project ARC DP210100227.

**Data Availability** All data are publicly available at https://genieclust.gagolewski.com and https://clustering-benchmarks.gagolewski.com.

## Declarations

**Conflict of Interest** The author declares no competing interests.

## References

Ackerman, M., Ben-David, S., Brânzei, S., & Loker, D. (2021). Weighted clustering: Towards solving the user's dilemma. *Pattern Recognition, 120*, 108152. https://doi.org/10.1016/j.patcog.2021.108152

Andrews, J., Browne, R., & Hvingelby, C. (2022). On assessments of agreement between fuzzy partitions. *Journal of Classification, 39*, 326–342.

Arbelaitz, O., Gurrutxaga, I., Muguerza, J., Pérez, J. M., & Perona, I. (2013). An extensive comparative study of cluster validity indices. *Pattern Recognition, 46*(1), 243–256. https://doi.org/10.1016/j.patcog.2012.07.021

Arinik, N., Labatut, V., & Figueiredo, R. (2021). Characterizing and comparing external measures for the assessment of cluster analysis and community detection. *IEEE Access, 9*, 20255–20276. https://doi.org/10.1109/ACCESS.2021.3054621

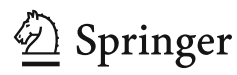

Arnold, B. C. (2015). *Pareto distributions*. New York, USA: Chapman and Hall/CRC. https://doi.org/10.1201/b18141

Braun-Blanquet, J. (1932). *Plant sociology*. The study of plant communities: McGraw-Hill.

Bullen, P. (2003). *Handbook of means and their inequalities*. Dordrecht: Springer Science+Business Media.

Caliński, T., & Harabasz, J. (1974). A dendrite method for cluster analysis. *Communications in Statistics, 3*(1), 1–27. https://doi.org/10.1080/03610927408827101

Campagner, A., Ciucci, D., & Denoeux, T. (2023). A general framework for evaluating and comparing soft clusterings. *Information Sciences, 623*, 70–93. https://doi.org/10.1016/j.ins.2022.11.114

Chacón, J. (2021). A close-up comparison of the misclassification error distance and the adjusted Rand index for external clustering evaluation. *British Journal of Mathematical and Statistical Psychology, 74*, 203–231.

Chacón, J., & Rastrojo, A. (2023). Minimum adjusted Rand index for two clusterings of a given size. *Advances in Data Analysis and Classification, 17*, 125–133.

Charon, I., Denoeud, L., Guénoche, A., & Hudry, O. (2006). Maximum transfer distance between partitions. *Journal of Classification, 23*, 103–121.

Crouse, D. (2016). On implementing 2D rectangular assignment algorithms. *IEEE Transactions on Aerospace and Electronic Systems, 52*(4), 1679–1696. https://doi.org/10.1109/TAES.2016.140952

Dasgupta, S., & Ng, V. (2009). Single data, multiple clusterings. Proc. NIPS workshop: Clustering: Science or art? Towards principled approaches. Retrieved from https://clusteringtheory.org

Dua, D., & Graff, C. (2022). UCI Machine Learning Repository. Irvine, CA. (http://archive.ics.uci.edu/ml)

Dunn, J. (1974). A fuzzy relative of the ISODATA process and its use in detecting compact well-separated clusters. *Journal of Cybernetics, 3*(3), 32–57. https://doi.org/10.1080/01969727308546046

D'Ambrosio, A., Amodio, S., Iorio, C., Pandolfo, G., & Siciliano, R. (2021). Adjusted Concordance Index: An extension of the adjusted Rand index to fuzzy partitions. *Journal of Classification, 38*, 112–128.

Flynt, A., Dean, N., & Nugent, R. (2019). sARI: A soft agreement measure for class partitions incorporating assignment probabilities. *Advances in Data Analysis and Classification, 13*, 303–323.

Fowlkes, E., & Mallows, C. (1983). A method for comparing two hierarchical clusterings. *Journal of the American Statistical Association, 78*(383), 553–569.

Fränti, P., & Sieranoja, S. (2018). K-means properties on six clustering benchmark datasets. *Applied Intelligence, 48*(12), 4743–4759.

Gagolewski, M. (2021). genieclust: Fast and robust hierarchical clustering. SoftwareX, 15, 100722. Retrieved from https://genieclust.gagolewski.com, https://doi.org/10.1016/j.softx.2021.100722

Gagolewski, M. (2022). A framework for benchmarking clustering algorithms. SoftwareX, 20, 101270. Retrieved from https://clustering-benchmarks.gagolewski.com, https://doi.org/10.1016/j.softx.2022.101270

Gagolewski, M., Bartoszuk, M., & Cena, A. (2021). Are cluster validity measures (in)valid? *Information Sciences, 581*, 620–636. https://doi.org/10.1016/j.ins.2021.10.004

Gagolewski, M., et al. (2022). A benchmark suite for clustering algorithms: Version 1.1.0. Retrieved from https://github.com/gagolews/clustering-data-v1/releases/tag/v1.1.0, https://doi.org/10.5281/zenodo.7088171

Gates, A., & Ahn, Y.-Y. (2017). The impact of random models on clustering similarity. *Journal of Machine Learning Research, 18*(87), 1–28.

Gates, A., Wood, I., Hetrick, W., & Ahn, Y. (2019). Element-centric clustering comparison unifies overlaps and hierarchy. *Scientific Reports, 9*(1), 8574. https://doi.org/10.1038/s41598-019-44892-y

Goodman, L., & Kruskal, W. (1979). *Measures of association for cross classifications*. Springer-Verlag.

Grabisch, M., Marichal, J.-L., Mesiar, R., & Pap, E. (2009). *Aggregation functions*. Cambridge University Press.

Graves, D., & Pedrycz, W. (2010). Kernel-based fuzzy clustering: A comparative experimental study. *Fuzzy Sets and Systems, 161*, 522–543.

Halkidi, M., Batistakis, Y., & Vazirgiannis, M. (2001). On clustering validation techniques. *Journal of Intelligent Information Systems, 17*, 107–145. https://doi.org/10.1023/A:1012801612483

Hennig, C. (2015). What are the true clusters? *Pattern Recognition Letters, 64*, 53–62. https://doi.org/10.1016/j.patrec.2015.04.009

Horibe, Y. (1985). Entropy and correlation, SMC-15(5), 641–642. https://doi.org/10.1109/TSMC.1985.6313441

Horta, D., & Campello, R. (2015). Comparing hard and overlapping clusterings. *Journal of Machine Learning Research, 16*(93), 2949–2997.

Hubert, L., & Arabie, P. (1985). Comparing partitions. *Journal of Classification, 2*(1), 193–218. https://doi.org/10.1007/BF01908075

Hüllermeier, E., Rifqi, M., Henzgen, S., & Senge, R. (2012). Comparing fuzzy partitions: A generalization of the Rand index and related measures. *IEEE Transactions on Fuzzy Systems, 20*(3), 546–556. https://doi.org/10.1109/TFUZZ.2011.2179303

Kvalseth, T. (1987). Entropy and correlation: Some comments. *IEEE Transactions on Systems, Man, and Cybernetics, 17*(3), 517–519. https://doi.org/10.1109/TSMC.1987.4309069

Lei, Y., Bezdek, J., Romano, S., Vinh, N., Chan, J., & Bailey, J. (2017). Ground truth bias in external cluster validity indices. *Pattern Recognition, 65*, 58–70. https://doi.org/10.1016/j.patcog.2016.12.003

Luna-Romera, J., Ballesteros, M., García-Gutiérrez, J., & Riquelme, J. (2019). External clustering validity index based on chi-squared statistical test. *Information Sciences, 487*, 1–17. https://doi.org/10.1016/j.ins.2019.02.046

Maulik, U., & Bandyopadhyay, S. (2002). Performance evaluation of some clustering algorithms and validity indices. *IEEE Transactions on Pattern Analysis and Machine Intelligence, 24*(12), 1650–1654. https://doi.org/10.1109/TPAMI.2002.1114856

Meilă, M. (2005). Comparing clusterings - an axiomatic view. S. Wrobel & L. De Raedt (Eds.), Proc. Intl. Machine Learning Conference (ICML) (pp. 577-584). https://doi.org/10.1145/1102351.1102424

Meilă, M., & Heckerman, D. (2001). An experimental comparison of modelbased clustering methods. *Machine Learning, 42*, 9–29. https://doi.org/10.1023/A:1007648401407

Milligan, G. W., & Cooper, M. C. (1985). An examination of procedures for determining the number of clusters in a data set. *Psychometrika, 50*(2), 159–179.

Morey, L., & Agresti, A. (1984). The measurement of classification agreement: An adjustment to the Rand statistic for chance agreement. *Educational and Psychological Measurement, 44*(1), 33–37. https://doi.org/10.1177/0013164484441003

Müllner, D. (2011). Modern hierarchical, agglomerative clustering algorithms. ArXiv:1109.2378.

Pedregosa, F., et al. (2011). Scikit-learn: Machine learning in Python. *Journal of Machine Learning Research, 12*(85), 2825–2830. Retrieved from http://jmlr.org/papers/v12/pedregosa11a.html

Rand, W. (1971). Objective criteria for the evaluation of clustering methods. *Journal of the American Statistical Association, 66*(336), 846–850. https://doi.org/10.2307/2284239

Rezaei, M., & Fränti, P. (2016). Set-matching measures for external cluster validity. *IEEE Transactions on Knowledge and Data Engineering, 28*(8), 2173–2186. https://doi.org/10.1109/TKDE.2016.2551240

Rousseeuw, P. J. (1987). Silhouettes: A graphical aid to the interpretation and validation of cluster analysis. *Journal of Computational and Applied Mathematics, 20*, 53–65. https://doi.org/10.1016/0377-0427(87)90125-7

Steinley, D. (2004). Properties of the Hubert-Arabie adjusted Rand index. *Psychological Methods, 9*(3), 386–396. https://doi.org/10.1037/1082-989X.9.3.386

Strobl, C., & Leisch, F. (2022). Against the "one method fits all data sets" philosophy for comparison studies in methodological research. *Biometrical Journal*. https://doi.org/10.1002/bimj.2022001042

Tavakkol, B., Choi, J., Jeong, M., & Albin, S. (2022). Object-based cluster validation with densities. *Pattern Recognition, 121*, 108223. https://doi.org/10.1016/j.patcog.2021.108223

Thrun, M., & Ultsch, A. (2020). Clustering benchmark datasets exploiting the fundamental clustering problems. *Data in Brief, 30*, 105501. https://doi.org/10.1016/j.dib.2020.105501

Ullmann, T., Hennig, C., & Boulesteix, A.-L. (2022). Validation of cluster analysis results on validation data: A systematic framework. *Wiley Interdisciplinary Reviews: Data Mining and Knowledge Discovery, 12*(3), e1444. https://doi.org/10.1002/widm.1444

van der Hoef, H., & Warrens, M. (2019). Understanding information theoretic measures for comparing clusterings. *Behaviormetrika, 46*, 353–370. https://doi.org/10.1007/s41237-018-0075-7

van Mechelen, I., Boulesteix, A.-L., Dangl, R., et al. (2023). A white paper on good research practices in benchmarking: The case of cluster analysis. Wiley Interdisciplinary Reviews: Data Mining and Knowledge Discovery, e1511. https://doi.org/10.1002/widm.1511

Vinh, N., Epps, J., & Bailey, J. (2010). Information theoretic measures for clusterings comparison: Variants, properties, normalization and correction for chance. *Journal of Machine Learning Research, 11*, 2837–2854.

von Luxburg, U., Williamson, R., & Guyon, I. (2012). Clustering: Science or art? I. Guyon et al. (Eds.), Proc. ICML Workshop on Unsupervised and Transfer Learning (Vol. 27, pp. 65-79).

Wagner, S., & Wagner, D. (2006). Comparing clusterings - an overview (Tech. Rep. No. 2006-04). Faculty of Informatics, Universität Karlsruhe (TH). Retrieved from https://i11www.iti.kit.edu/extra/publications/ww-cco-06.pdf

Warrens, M., & van der Hoef, H. (2022). Understanding the adjusted Rand index and other partition comparison indices based on counting object pairs. *Journal of Classification, 39*, 387–509. https://doi.org/10.1007/s00357-022-09413-z

Xiang, Q., et al. (2012). A split-merge framework for comparing clusterings. Proc. Intl. Machine Learning Conference (ICML) (pp. 1259-1266).
Xiong, H., & Li, Z. (2014). Clustering validation measures. C. Aggarwal & C. Reddy (Eds.), Data clustering: Algorithms and applications (pp. 571-606). CRC Press.
Xu, Q., Zhang, Q., Liu, J., & Luo, B. (2020). Efficient synthetical clustering validity indexes for hierarchical clustering. *Expert Systems with Applications, 151*, 113367. https://doi.org/10.1016/j.eswa.2020.113367